\documentclass[12pt]{article}
\usepackage[utf8]{inputenc}
\usepackage{geometry}
\usepackage{color}
\usepackage[ruled,vlined]{algorithm2e}
\usepackage{amsmath}
\usepackage{amssymb}
\usepackage{booktabs}
\usepackage{graphicx}
\usepackage{setspace}
\usepackage[authoryear]{natbib}
\usepackage[unicode=true,
 bookmarks=false,
 breaklinks=false,pdfborder={0 0 1},backref=false,colorlinks=true]
 {hyperref}
\hypersetup{breaklinks,hyperindex,citecolor=blue,linkcolor=blue}

\makeatletter
\@ifundefined{date}{}{\date{}}
\usepackage{pdfcomment}
\usepackage{environ}
\usepackage{authblk}
\usepackage{amsthm}\usepackage{multirow}
\usepackage{multicol}\usepackage{mathrsfs}
\usepackage{color}
\usepackage{caption}
\usepackage{subcaption}
\usepackage{url}
\usepackage{tikz}
\usetikzlibrary{external}
\usepackage{pgfplots}
\usepackage{pdflscape}
\usepackage{eqparbox}
\usepackage{lipsum}

\bibpunct{(}{)}{;}{a}{}{,}

\graphicspath{{DFSFigures/}}

\newtheorem{theorem}{Theorem}

\newtheorem{proposition}{Proposition}
\newtheorem{corollary}[theorem]{Corollary}

\newtheorem{remark}{Remark}
\newcommand{\btheta}{\boldsymbol{\theta}}

\newcommand{\bSigma}{\boldsymbol{\Sigma}}

\newcommand{\balpha}{\boldsymbol{\alpha}}

\newcommand{\bmu}{\boldsymbol{\mu}}

\newcommand{\B}{\mathbf{B}}

\newcommand{\bA}{\mathbf{A}}

\newcommand{\bE}{\mathbf{E}}
\newcommand{\bI}{\mathbf{I}}
\newcommand{\X}{\mathbf{X}}

\newcommand{\s}{\mathbf{s}}

\newcommand{\br}{\mathbf{r}}
\newcommand{\bb}{\mathbf{b}}

\newcommand{\calR}{\mathcal{R}}

\newcommand{\calA}{\mathcal{A}}

\DeclareMathOperator{\diag}{diag}

\makeatother
\author{\small{Jie Guo$^1$,   Hao Yan$^2$,   Chen Zhang$^{1}$\footnote{Corresponding author},   Steven Hoi$^3$}}

\begin{document}
	\title{Partially Observable Online Change Detection via Smooth-Sparse Decomposition}
	\date{}
	\maketitle
\begin{center}{\small{$^1$Tsinghua University}}
\par\end{center}
\begin{center}{\small{$^2$Arizona State University}}
\par\end{center}
\begin{center}{\small{$^3$Singapore Management University}}
\par\end{center}
	\begin{abstract}
We consider online change detection of high dimensional data streams with sparse changes, where only a subset of data streams can be observed at each sensing time point due to limited sensing capacities. On the one hand, the detection scheme should be able to deal with partially observable data and meanwhile have efficient detection power for sparse changes. On the other, the scheme should be able to adaptively and actively select the most important variables to observe to maximize the detection power. To address these two points, in this paper, we propose a novel detection scheme called CDSSD. In particular, it describes the structure of high dimensional data with sparse changes by smooth-sparse decomposition, whose parameters can be learned via spike-slab variational Bayesian inference. Then the posterior Bayes factor, which incorporates the learned parameters and sparse change information, is formulated as a detection statistic. Finally, by formulating the statistic as the reward of a combinatorial multi-armed bandit problem, an adaptive sampling strategy based on Thompson sampling is proposed. The efficacy and applicability of our method in practice are demonstrated with numerical studies and a real case study.  
	\end{abstract}

\section{Introduction}
%However, processing these big data streams in real time is really a big challenge. With some constrains in real world, we can't handle the collection, transmission or analysis of these big volume and high dimensional data. First, collecting big data streams needs a large number of sensors, which places a heavy burden on the cost of sensor purchasing and allocating. Second, the life time of sensors' battery is limited whereas the change of this batch of sensors is inconvenient and costly. Thus to save the energy of sensors' battery, only a subset of sensors can be set "ON" all the time. Futhermore, the transmisson bandwidth to transmit these large volume data, the memory and computation speed of computers to analyse these large volume data are limited. All these limitations can result that only a subset of variables in the system can be observed at every time epoch. Take the solar flare detection problem as an example, which will be dicussed more in case study later. Pictures of the surface of sun are shot with high frequency by cameras in the space station. The brightness of every pixels in one image is treated as a variable, so thousands of the pixels in one HD image are the variables we observe. However, we can't transmit so huge data streams back to the fucsion center on Earth due to the limited transmission bandwidth. An effective solution to this problem is to adaptively sample a subset of all pixels to tansmit and monitor. To track and detect the anomaly in time, naturally, we should transmit those variables with high probability to shift from those normal states.
High dimensional sequential change point detection has been extensively studied in statistics and machine learning. Sequential samples from $p$ variables, $\X_{1}, \X_{2},\ldots$ are identically and independently distributed from a distribution in a dimensional space $\X \in \calR^{p}$. The $p$ variables of each sample may have complex correlations with each other, depending on the data structure of $\X_{t}$. For example, $\X_{t}$ can be a vector, a profile, an image, etc. We aim at detecting a possible change point $\tau$. Before it, the samples $\X_{t}, t\leq \tau$ follow a known distribution $f_0$. After the change point, the samples $\X_{t}$ follow another unknown post-change distribution $f_{1}$. The goal is to detect the unknown change point as soon as possible after it occurs. We restrict our attention to detecting one change point, which often arises in sequential monitoring problems \citep{montgomery2007introduction}.

For high dimensional data modeling, the correlation information of variables is hard to compute due to the curse of dimensionality. Consequently, dimension reduction methods, such as dictionary learning and matrix decomposition methods \citep{cheng2018tensor,qi2017multi}, are usually adopted to describe the high dimensional data in the feature level. Furthermore, when a change happens, it usually affects a few variables simultaneously. So we need to consider the variable correlation structure existing in the change as well. In other words, the dictionary should include both patterns of normal data and the patterns of the changed(abnormal) data. We further assume that when a change happens, it can be linearly represented by a few anomaly patterns in the dictionary\citep{mo2013adaptive}. Considering the dictionary is large, the linear representation would be sparse.  This brings new demand for more powerful detection schemes. Specifically, if we know that which potential abnormal patterns would affect which variables, then the corresponding detection scheme will only focus on these patterns and filter out background noise from other unchanged features.  

Besides the challenges above, another emerging challenge in sequential change detection is limited sensing resources. 
%In reality, to achieve quality control of an industrial or a service system, there are sensors collecting the state information of variables in real time. With the rapid development of sensor technology, the number of sensors distributed to collect state information of index variables in a system is increasing rapidly and forms big data streams. Since these data streams may involve anomaly information of some variables, utilizing and analysing these rich information thoroughly can help us distinguish the anomaly and improve the system's performance with a great deal. 
Classical researches for high dimensional data sequential sparse change detection focus on a fully observable process, i.e., at each sampling time point, all the $p$ variables in $\X_{t}$ can be observed for analysis. However, in reality, sometimes it is unfeasible to acquire measurements of all these variables in real time. Instead, only a subset of the $p$ variables can be accessible, such as in the following scenarios: (1) when the number of sensors cannot exceed certain number due the limited sensing resources; (2) when only a limited number of sensors can be set at "ON" mode due to limited battery lifetime; (3) when only partial data collected at each acquisition time can be transmitted back to fusion center for real-time analysis due to limited transmission bandwidth and computing speed. In any of the above-mentioned scenarios, only a subset of $\X_{t}$ with $m$ variables out of the $p$ variables ($m \leq p$) can be observed. This further increases the difficulty of high dimensional sparse change detection. It requires us to not only deal with partial observations, but also develop a smart sensor allocation strategy to choose which variables to observe at each time point. Otherwise, if variables containing change patterns can not be observed, the change would never be detected.

In this paper, we aim at this problem of Partially Observable High Dimensional data Sequential Sparse Change Detection (POHDSSCD). Our goal is to develop a sequential detection algorithm for sparse changes, which can dynamically choose a subset of variables to observe at each time point such that the detection power can be maximized without violating the sensing constraints. 
%Also, we need to balance between the exploration and exploitation. On one hand, we need to explore all the sensing variables to localize the defect. On the other hand, we need to focus our sensing power on the suspected areas for the change. 
In particular, (i) we describe the structure of high dimensional correlated data streams with sparse changes in feature level by smooth-sparse decomposition (SSD) \citep{yan2017anomaly}. The decomposition, on one hand, can describe the data feature for before-change distribution $f_0$, and on the other, can be customized to detect any specified sparse change $f_1$. (ii) Under the Bayesian learning framework with partial observations, we use the spike-slab variational Bayesian inference to learn the parameters of the decomposition, based on which a detection statistic is constructed by the posterior Bayes factor. (iii) Furthermore, we formulate the detection statistic as the reward function in the combinatorial multi-armed bandit problem and propose a Thompson sampling strategy to decide the most informative subset of variables to observe for the next sampling point such that the detection power can be maximized. 

The remainder of the article is organized as follows. In Section \ref{sec:literature}, we review the literature of some related topics to the proposed problem. Section \ref{sec:formulation} describes more specific problem formulation. Section \ref{sec:method} introduces the main body of our proposed method sequentially, including variational Bayesian inference, Bayesian hypothesis testing and Thompson sampling. Section \ref{sec:simulation} presents a simulation study on synthetic data and a real-world case study to further illustrate the efficacy and effectiveness of the proposed method.
	
\section{Related Works}
\label{sec:literature}
To better describe the proposed framework, we would like to discuss some related state-of-the-art methods in the field of statistics and machine learning.

\textbf{Sequential sparse change detection} for multivariate streaming data has recently attracted increasing attention in many applications. Considering for the high dimensional data where only a sparse subset of variables may be affected by the change, many works utilized the idea of sparse learning for sparse change detection \citep{chan2017optimal,wang2013montoring}. For example, \cite{wang2009high} proposed a penalized likelihood function to screen out potential out-of-control variables. Similarly, \cite{zou2009multivariate} adopted LASSO regularity to force sparse regularization on the estimated changes. %\cite{mei2010efficient} proposed a top-R method by only selecting the most likely changed variables into the final monitoring scheme. 
%multivirate cumulative sum () global online monitoring schemes that are scalable and robust when the number of affected data streams is either moderately large or completely unknown.
%Considering the sparsity of affected variables in an anomaly, 
% Then \cite{chan2017optimal,wang2013montoring}  proposed several change detection schemes for high dimensional cases by considering the sparsity of the change. 
Most of these methods assume the correlation matrix of different variables is known in advance or can be estimated via some historical data. Yet this is not true for the high-dimensional process due to the curse of dimensionality. To solve it, one kind of method is to modify the estimation of the correlation matrix, by assuming it is diagonal \citep{mei2010efficient}. Then univariate detection statistics for each dimension are constructed separately, but only statistics of the top R most likely changed variables are fused together as the final statistic to filter out noises \citep{mei2010efficient}. However, this loss of correlation information compromises the detection power a lot. Another kind of method is to use dimension reduction or a low-rank approximation to describe the correlation structure. In particular, \cite{zhang2018weakly} proposed a sparse functional principal component analysis (PCA) to model multi-channel profiles. Then the sparse PCA scores are used to construct a detection statistic for online monitoring multi-channel profile data. \cite{yan2017anomaly} described the high dimensional spatial correlation in image data by smooth-sparse decomposition. Then the sparse anomalous regions are learned and the LASSO based detection statistic of \cite{zou2009multivariate} is constructed. However, all these methods can only be applied in fully observable scenarios, therefore do not apply for partially observed data.
% considered the weakly correlation and sparse change in the monitoring of multi-channel profile data, and proposed a sparse multi-channel functional principal component analysis to model multi-channel profiles and added the LASSO penalty on PCA scores so that each profile is forced to be a sparse combination of a small subset of the extracted features.\cite{yan2017anomaly} described the high dimension and complex spatial correlation in image data by smooth-sparse decomposition, which decomposes an image into three components, the smooth image background, the sparse anomalous regions and the random noises. This decomposition helped with the anomaly detection accuracy and computation efficiency for image data. However, all these methods can be only applied in fully observable scenarios, and cannot be applied in our POHDSSCD.

\textbf{Partial observable sequential change detection} is an emerging topic that has not been fully addressed. The most pioneer work \cite{liu2015adaptive} proposed a top-R detection scheme by extending \cite{mei2010efficient} to the scenario with missing observation. Later \cite{xian2017nonparametric} extended the work of \cite{liu2015adaptive} to non-Gaussian process, by constructing an anti-rank detection statistic based on data spatial structure. However, these two methods treat different variables as independent without exploiting their correlation structure. This leads to their methods perform poorly in some scenarios, as shown in Section \ref{sec:simulation}. On the one hand, taking advantage of the correlation structure can improve the detection efficiency, especially when the change influences some sensors jointly. On the other hand, if a variable is not observed, its information can still be inferred based on its correlation with other observed variables. Later \cite{wang2018spatial} proposed a spatial-adaptive sampling and monitoring procedure that utilized the spatial information of the data streams for quick change detection. \cite{xian2019online} revised the rank-based statistic of \cite{xian2017nonparametric} by containing the correlation information. It can automatically augment information for unobservable variables based on other observed ones, and intelligently allocate the monitoring resources to the most suspicious data streams. However, all these adaptive sampling strategies are heuristic, and their adaptive sampling strategies are based on rule of thumb without any theoretical guarantee. Recently, \cite{zhang2019partially} exploited the relationship between partial observable online detection with a combinatorial multi-armed bandit, and proposed an adaptive sampling strategy based on the upper confidence bound (UCB) algorithm. This work also analyzed the theoretical property of lower bound of detection power. However, it has the limitation of huge computation complexity and is unpractical to be applied in the high dimensional process. 
%the number of selected variables $m$ or the number of full variables $p$ is large, searching all the sets $M=\tbinom{p}{m}$ is time consuming, so the algorithm is impractical for large-scale problems
In addition, to deal with the correlation of variables, all existing methods assume the covariance matrix is known and directly use it to formulate the monitoring statistic. As mentioned earlier, this cannot be satisfied in reality. Furthermore, these methods do not target at sparse change, and consequently have limited power for the POHDSSCD problem.  

\textbf{Multi-armed bandit} (MAB) is a problem extensively studied in reinforcement learning and online learning. It considers a system with $p$ arms where in each round one arm (or a combinatorial subset of arms) can be selected and a reward is achieved. The reward of each arm (or each combinatorial set) follows a certain distribution with unknown expectation, and the objective of MAB (or combinatorial MAB, i.e., CMAB) is to play these arms in sequential rounds with an arm selection policy such that the total expected reward can be maximized. 
%However, for combinatorial multi-armed bandit (CMAB) problem \citep{chen2013combinatorial}, the arms played each time is not a simple arm, but a combination of multiple arms, and the reward generated by these arms may include non-linear functions. 
In our scenario, %based on historical observations and the results of parameter inference, 
we want to sequentially decide the best subset of variables so as to minimize the average detection delay, which is similar to the objective of CMAB. Hence we can borrow some ideas from MAB related works. So far, a number of studies have been done on developing sampling strategies for MAB problems. In general, they can be classified into two categories: the UCB \citep{chen2013combinatorial} and the Thompson sampling algorithms \citep{durand2014thompson}. 
Built upon them, some works also have studied how to choose the best top-$K$ arms \citep{even2006action,bubeck2013multiple}, or the outlier arms \citep{zhuang2017identifying}. However, these methods assume the system is static, i.e., the reward distribution of arms do not change sequentially. Yet our problem is more about identifying the change of the system. Furthermore, they target at single or top-$K$ arms (or variables) identification. Yet in our problem, we would like focus on system level detection. 
%One solution to CMAB is combinatoral upper confidence bounded (CUCB) algorithm \citep{chen2013combinatorial}. It introduces an upper bound of super arms' reward, which can be tuned with the observation times to balance the exploration and the exploitation. The other one is Thompson sampling which is implemented by sampling a value of related parameter from its distribution and selecting a subset of arms that maximize the reward function to achieve both exploration and exploitation \citep{durand2014thompson}. 
Recently, there are also some works combining change point detection algorithms with bandit algorithms. Considering that the reward of each arm is not stationary but piecewise-constant and the shifts at unknown time points are change points, \cite{liu2018change,cao2019nearly} combined change point detection procedures with UCB method, to track the time-varying reward distributions. Yet their objective is still maximizing the total expected reward, instead of system change detection. 
\section{Problem Formulation}
\label{sec:formulation}
Consider a system consisting of $p$ variables. Denote the signals of these variables at sensing time point $t$ as $\mathbf{X}_{t} = ( X_{1t},\ldots,X_{pt})$. We assume in the normal condition $\mathbf{X}_t \stackrel{iid}{\sim} f_0$ for $t=1,2,\ldots$, and we are interested in detecting any change of these $p$ variables. For high-dimensional $\mathbf{X}_t$, dictionary learning and representation is commonly used to reduce dimension and describe the complex correlation structure of data \citep{cheng2018tensor,qi2017multi}.
%\citep{zhao2016spectral,mairal2009online,meier2009high}. The decomposition method has been widely adopted to explain many kinds of complex data, such as high-dimensional data \citep{meier2009high}, tensor data \citep{anandkumar2014tensor}, functional data \citep{ramsay2004functional}, spatio-temporal data \citep{yu2016temporal}, graphical data \citep{cai2010graph}, etc. 
Following their general decomposition formulation, we assume for normal $\X_{t}$, it can be expanded on a before-change feature space with $k_{b}$ bases $\B_{b}=[\bb_{b1},\ldots,\bb_{b k_b}]\in \calR^{p \times k_{b}}$, i.e.,
\begin{align}
\label{eq:IC_decomposition}
\X_{t} = \B_{b} \btheta_{t}+\mathbf{E}_{t},
\end{align} 
where $\btheta_{t}\in \calR^{k_b \times 1}$ are the coefficients and $\bE_{t}\in \calR^{p \times 1}$ are the noise terms. In this paper, we assume $\bE_{t}$ follows a Gaussian distribution as $\bE_{t}\sim N(\mathbf{0},\bSigma_{e})$ with $\bSigma_{e}=\sigma_{e}^{2}\bI$. Here $\B_{b}$ can be either learned by historical observed samples via matrix decomposition algorithms, or be set as notable spaces such as Spline space \citep{meier2009high}, Fourier space, Kernel space, etc \citep{wang2018atomo}. Consequently, the decomposition can explain the data covariance matrix as $\mathrm{Cov}(\X_{t})=\B_{b}\mathrm{Cov}(\btheta_{t})\B_{b}^{\prime}+\bSigma_{e}$. Since the rank of $\mathrm{Cov}(\btheta_{t})$ is generally much smaller than that of $\X_{t}$, this is exactly the low rank estimation for high-dimensional covariance matrix \citep{fan2008high,cai2016estimating}.
%and has been generally used in the literature \citep{fan2008high,cai2016estimating}, in which the covariance matrix can be represented as the sum of a low-rank matrix of multi-factors and a sparse matrix of uncorrelated errors.
In this paper, without loss of generality, we further assume the projection of $\X_{t}$ on the before-change bases is stable and $\btheta_{t}$ follows a Gaussian distribution with mean $\mathbf{0}$ and covariance matrix $\bSigma_{0}$. Then we can have $\mathrm{Cov}(\X_{t})=\B_{b}\bSigma_{0}\B_{b}^{\prime}+\bSigma_{e}$. 

When $\X_{t}$ occurs sparse changes, unlike the before-change distribution of $f_0$ focusing on low-rank structures, many types of sparse changes with diverse anomaly patterns may occur in the system, and the chance that each type of anomaly pattern happens is pretty small. With this in mind, we may further define an anomaly dictionary with a set of anomaly bases $\B_{a}=[\bb_{a1},\ldots,\bb_{a k_a}]$.
%which includes any possible sparse change pattern (i.e., $\bb_{aj},j=1,\ldots,k_{a}$ are sparse vectors). 
Here $k_a$ can be even larger than $p$. $\B_{a}$ can either be set by domain knowledge from practitioners, if certain specific change patterns are of interest, or be learned from some collected anomaly data via the dictionary learning approach. In conclusion, we can utilize a composite decomposition approach to describe the post-change distribution $f_1$ with sparse change patterns from $f_0$, i.e.,
%In particular, similar to expanding $\X_{t}$ on before-change bases as the before-change $\X_{t}, t \leq \tau$ does, we can decompose the post-change $\X_{t}$, in the way as
\begin{align}
\label{eq:OC_decomposition}
\X_{t} = \B_{b}\btheta_{t}+\B_{a}\btheta_{a} + \mathbf{E}_{t}. \quad \forall t>\tau
\end{align}
In this paper, 
% without loss of generality, we assume $\B_{b}$ and $\B_{a}$ are well defined orthogonal bases with $\B_{b}^{\prime}\B_{b} = \I$ and $\B_{a}^{\prime}\B_{a} = \I$. 
considering that in most of the applications, the anomaly bases generally have components outside the subspace spanned by the normal bases \citep{xu2020factorized,zhang2018supervised}, we further assume that $\B_{b}^{\prime}\B_a= \mathbf{0}$.  

This concept of composite decomposition can be dated back to additive models \citep{wood2015generalized}, where a nonparametric regression is defined as a combination of several composite models. Later similar concepts have been applied in many applications. \cite{ba2012composite} proposed a ``composite Gaussian process'' to describe global features and local features of expensive functions. %The first Gaussian process is required to be smoother to capture the global trend, while the second Guassian process models local adjustments.
\cite{zhang2016monitoring} constructed an additive Gaussian process model with two separate Gaussian processes to describe characteristics of desired profile and abnormal profile data. \cite{yan2017anomaly} decomposed image signal into a smooth functional mean plus sparse anomalous regions, for image anomaly detection. 
%By introducing a penalized nonparametric regression model for SSD and proposing efficient algorithms, they developed a fast optimization algorithm to solve model parameters. Later, \citep{yan2018real} extended this idea to temporal dimension and modeled both the temporal trend and the spatial structure of image signals through spatio-temporal decomposition approach.

Combining (\ref{eq:IC_decomposition}) and (\ref{eq:OC_decomposition}), we can define the change-point model as this: $\btheta_{a}=\mathbf{0}, \forall t\leq \tau$, and $\btheta_{a}\neq \mathbf{0}, \forall t > \tau$. In real online change detection scenario, since $\tau$ is unknown, our goal is to construct a hypothesis test to decide whether 
\begin{align}
\label{eq:hypothesistest}
H_{0}:\ \btheta_{a} = \mathbf{0}, \quad H_{1}:\  \btheta_{a} \neq \mathbf{0},
\end{align}
for each time point $t$, based on the partially observed subset of $\X_{t}$. %The test statistic should be able to detect the system change as soon as possible, with small detection delay. 
Here we introduce a sensing variable $z_{it}$ for each variable $X_{it}$ such that $z_{it} = 1$ if and only if $\ X_{it}$ is observed at time point $t$, and the sensing constraint can be expressed as $\sum_{i = 1}^{p}, z_{it} = m, \forall t$. Denote $Z(t)$ to be the vector of indices corresponding to the observed dimensions for $\X_{t}$. $\X_{Z(t)}\in \calR^{m \times 1}$ represents the observed data for time point $t$. 

We would like construct a detection scheme for (\ref{eq:hypothesistest}). We assume that after $\tau$, the change would keep. We construct the scheme relating to a stopping time $T$ associated with a test statistic $\Lambda(t)$.
%A general detection scheme to decide whether a change occurs is related to a stopping time $T$ associated with a test statistic $\Lambda(t)$, $\forall t$. 
The scheme defines a stopping time $T=\inf_{t}\{\Lambda(t)>h\}$ where $h$ is a pre-defined constant threshold, and $T=n$ is explained as the detection scheme stops at time $n$ and indicates that there exists a change among the first $n$ time points. The performance of the detection scheme can be evaluated by two criteria: Average Run Length (ARL), before a false alarm occurs in normal condition, i.e., $ARL_{0}=E(T|\tau=\infty)$, and Average Detection Delay (ADD) after a change occurs in abnormal condition, i.e., $ADD_{\tau}=E(T-\tau|T>\tau, \tau < \infty)$. In practice, conditional on ARL as a fixed number which controls the false alarm rate, a detection scheme is formulated to minimize ADD. 
%the problem can be formulated as finding a change detection scheme $T$ that minimizes ADD with imposing a constraint on $ARL_{0}$, .
%Due to limited sensing resources, at each sensing epoch, we can only observe $m(m \leq p)$ out of $p$ variables.
\section{Our Method}
\label{sec:method}
In this section, we propose a change detection scheme for POHDSSCD in the Bayesian framework. 
%Change point detection with Bayesian methods has been generally learned in literature \cite{adams2007bayesian,tartakovsky2010state}. We have a prior information about the parameters which can be estimated from historical data or from the experience of practitioners.As time goes by, 
In particular, we construct an online variational Bayesian inference to estimate the posterior distribution of both $\btheta_t$ and $\btheta_{a}$ with sequential samples $\X_{Z(t)}$. The online estimation can make the best use of historical data and detect the nonzero $\btheta_{a}$ efficiently. Then the posterior distributions of $\btheta_t$ and $\btheta_{a}$ are used to construct a detection statistic for (\ref{eq:hypothesistest}) based on posterior Bayes factor. The test statistic can be treated as the reward distribution of a CMAB problem, and accordingly, a Thompson sampling framework to maximize the reward is proposed for selecting observations for the next time point. 
\subsection{Spike-Slab Model}
\label{subsec:VB}
With the prior information that $\btheta_{a}$ is a sparse vector, we consider the prior distribution that each component of $\btheta_{a}$ follows a spike-slab model independently. The spike-slab prior has been commonly used in many models for sparse vector estimation \citep{mitchell1988bayesian}. In particular, binary variables $\br = [r_{1},\ldots,r_{k_a}]$ are introduced to indicate whether $\theta_{aj}$ is nonzero. $r_{j}$ is a Bernoulli random trial governed by common success rate $p(r_{j}=1)=w_{j}$. If $r_{j}=0$, $p_{0}(\theta_{aj})$ follows the Gaussian distribution with zero mean and variance $v\sigma_{j}^{2}$, with $v\ll1$, e.g., the ``spike'', which demonstrates that the probability $p_0(\theta_{aj})=0$ almost equals $1$. Otherwise, $p_{0}(\theta_{aj})$ follows the Gaussian distribution with zero mean and variance $\sigma_{j}^{2}$, e.g., the "slab", which demonstrates that the probability $p_{0}(\theta_{aj} \neq 0)$ is large. This hierarchical prior distribution of $\btheta_a$ can be written as 
\begin{align}
\label{eq:ssprior}
p_{0}(\theta_{aj}|r_{j},\sigma_{j}^{2})& \sim N \left( 0, r_{j}\sigma_{j}^{2}+(1-r_{j})v\sigma_{j}^2 \right), \\ \nonumber
p_{0}(r_{j})& \sim Bernoulli (w_{j}), \ \ j = 1,\ldots,k_{a}. 
\end{align}
%where $r_{j}$ follows a Bernoulli distribution with $\mathrm{E}(r_{j})=\alpha_{j}$, and $\delta_{0}(\cdot)$ is the delta mass (or ``spike'') at zero. 	
Based on (\ref{eq:ssprior}), supposing the current time is $n$, we aim to estimate the posterior probability of $\btheta_{a}$ based on all the previous $n$ samples $\X_{Z(t)}, t=1,\ldots,n$. Consider that samples in recent time points are more likely to represent the current system state and can better detect the changes of the current system state than samples in the past time points. We would like to impose more weights on the current time points in the estimation. As such, we enforce time decayed weights $\lambda_{t}^{n},t=1,\ldots,n$ on the $n$ samples, in the sense that $\lambda_{1}^{n}<\lambda_{2}^{n}\ldots<\lambda_{n}^{n}$, and get the weighted posterior distribution of $\btheta_{a}$ as
\begin{align}
\label{eq:posterior}
p\left( \btheta_{a},\br|\X_{Z(1)},\ldots,\X_{Z(n)} \right) \propto p_{0}(\btheta_{a},\br)\prod_{t=1}^{n} p(\X_{Z(t)}|\btheta_{a},\br)^{\lambda_{t}^{n}}.
\end{align}
In this paper, we use the exponential decayed weights, i.e., $\lambda_{t}^{n}=\frac{\lambda(1-\lambda)^{n-t}}{1-(1-\lambda)^{n}}$ with a small positive value $\lambda\in (0,0.1]$. 

With the spike-slab model structure, we can reformulate our hypothesis of (\ref{eq:hypothesistest}) as 
\begin{equation}
H_{0}:\ \btheta_{a} = \mathbf{0},\quad H_{1}:\ \btheta_{a} \sim N(\bmu_{r},\mathbf{K}), 
\end{equation}
where $\bmu_{r}=\bmu_{a}\circ \br$, with $\circ$ representing the element-wise product. $\bmu_{a}=[\mu_{1},\cdots,\mu_{k_a}]^{\prime}$ is the estimated abnormal mean of the slab distribution of $\btheta_{a}$ and
%and we assume its set of all possible values is $\mathscr{R}$. 
$\mathbf{K}=\diag\Big(((1-r_1)v+r_1)s_1^2,((1-r_2)v+r_2)s_2^2,...,((1-r_{k_a})v+r_{k_a})s_{k_a}^2\Big)$ is the estimated covariance matrix of $\btheta_a$.  
%-------------------------------
\subsection{Variational Bayesian Inference}
\label{subsec:inference}
Unfortunately, (\ref{eq:posterior}) does not have a closed-form solution. So here we propose to approximately estimate (\ref{eq:posterior}) using variational methods, which have been popularly adopted in the literature. Variational methods can achieve high efficiency in computing the posterior distributions when the number of parameters to be estimate is relatively large \citep{attias2000variational,carbonetto2012scalable}. 
%and should be estimated via numerical methods. Variational methods have become popular in the context of inference problem. For example, to learn a graphical model, \cite{attias2000variational} presented variational Bayesian to facilitate analytical calculations of posterior distributions over the hidden variables, parameters and structures by drawing together variational ideas and Bayesian inference. In addition, \cite{carbonetto2012scalable} adopted the Bayesian approach to approximate the posterior distributions of coefficients in a regression model and retained useful features at a reduced cost. 
%In Bayesian analysis, the objective is to compute the posterior probabilities, which is inherently a high-dimensional integration problem and is intractable for most models. Variational Bayesian inference solves this problem by using an approximating joint distributions to approach the true distributions and converting this problem to an optimization problem. By minimizing the Kullback-Leibler divergence between these two distribution, the approximating distribution can matches the true distribution well. To make this approach viable for large problems, we introduce the famous "mean field" theory, which means that the simple conditional independence property is satisfied among the approximating distributions of variables. 
Here, the idea of variational Bayesian approach is to approximate (\ref{eq:posterior}) via another distribution $q(\btheta_{a},\br)=\prod_{j=1}^{k_a}q_{j}(\theta_{aj},r_{j})$, such that its Kullback-Leibler divergence from the true posterior distribution  (\ref{eq:posterior}) is minimized. This can be done by iteratively updating each $q_{j}(\theta_{aj},r_{j})$ sequentially with other $q_{k}(\theta_{ak},r_{k}),k \neq j$ fixed until convergence. 
Following \cite{carbonetto2012scalable}, we restrict $q_{j}(\theta_{aj},r_{j})$ to still have the form 
\begin{align}
\label{eq:approximating posterior}
q_{j}(\theta_{aj}|r_{j})& \sim N(\theta_{aj}|r_j\mu_{aj},r_{j}s_{j}^{2}+(1-r_{j})vs_{j}^{2}),\\ \nonumber
q_{j}(r_{j}) &\sim Bernoulli (\alpha_{j}).
\end{align} 
%where $r_{j}$ follows a Bernoulli distribution with $\mathrm{E}(r_{j})=\alpha_{j}$, and $\delta_{0}(\cdot)$ is the delta mass (or ``spike'') at zero. 	
Finding the best fully-factorized distribution $q_{j}(\theta_{aj},r_{j})=q_{j}(\theta_{aj}|r_{j})q_{j}(r_{j})$ indicates to find $\{\mu_{aj},s_{j}^{2},\alpha_{j}\}$ that minimize the Kullback-Leibler divergence. This is equivalent to maximizing the negative Kullback-Leibler divergence
\begin{align}
Z &= \int{q(\btheta_{a},\br)\ln\frac{p( \btheta_{a},\br,\X_{Z(1)},\ldots,\X_{Z(n)})}{q(\btheta_{a},\br)}d q(\btheta_{a},\br)}  \\ \nonumber
& = E_{q(\btheta_{a},\br)}[\ln p\left( \btheta_{a},\br,\X_{Z(1)},\ldots,\X_{Z(n)} \right)] - E_{q(\btheta_{a},\br)}[\ln q(\btheta_{a},\br)].
\end{align}
%where $q(\btheta_{a},\br)=\prod_{j=1}^{k_a}q_{j}(\theta_{aj},r_{j})$, and 
%\[p(\btheta_{a},\br,\X_{Z(1)},\ldots,\X_{Z(n)})=p_{0}(\btheta_{a},\br)\prod_{t=1}^{n} p(\X_{Z(t)}|\btheta_{a},\br)^{\lambda_{t}^{n}}.\]	
The coordinate descent updates for this optimization problem can be obtained by taking the partial derivatives of the negative Kullback-Leibler divergence, setting the partial derivatives to zero, and solving for the parameter $\mu_{aj},s_{j}^{2}$, and $\alpha_{j}$. This yields coordinate updates:
\begin{align}
\label{eq:update_a}
\mu_{aj}&=\frac{s_{j}^{2}}{\sigma_{e}^{2}}\times \Big(\sum_{t=1}^{n}\lambda_{t}^{n}\big((\X_{Z(t)}-\B_{b}\tilde{\btheta}_{n})^{\prime}\B_{ajZ(t)}+\sum_{k\neq j}\B_{ajZ(t)}^{\prime}\B_{akZ(t)}\alpha_{k}\mu_{ak}\big)\Big),\\
% \mu_{aj}&=\frac{\sum_{t=1}^{n}\frac{\lambda_t^n}{\sigma_e^2}(\tilde{\X}_{Z(t)}^{\prime}\B_{ajZ(t)}-\sum_{k \neq j}\B_{ajZ(t)}^{\prime}\B_{akZ(t)}\alpha_k\mu_{ak})}{\sum_{t=1}^{n}\frac{\lambda_t^n}{\sigma_e^2}\B_{ajZ(t)}^{\prime}\B_{ajZ(t)}+\frac{1}{\sigma_j^2}}\\
\label{eq:update_s2}
(s_{j}^{2})^{-1}&=\sum_{t=1}^{n}\frac{\lambda_{t}^{n}\B_{ajZ(t)}^{\prime}\B_{ajZ(t)}}{\sigma_{e}^{2}}+\frac{1}{\sigma_{j}^{2}},\\
% (s_j^2)^{-1}&=\sum_{t=1}^{n}\frac{\lambda_t^n}{\sigma_e^2}\B_{ajZ(t)}^{\prime}\B_{ajZ(t)}(1+v(\frac{1}{\alpha_j}-1))+\frac{1}{\sigma_j^2}\\
\label{eq:update_alpha}
% \ln \frac{\alpha_{j}}{1 - \alpha_{j}}&= \ln\frac{w_{j}}{1 - w_{j}}-\frac{{\mu_{aj}}^{2}}{2{\sigma_{j}}^{2}}+\frac{{\mu_{aj}}^{2}}{{s_{j}}^{2}}- \sum_{t=1}^{n}\frac{\lambda_{t}^{n}\B_{ajZ(t)}^{\prime}\B_{ajZ(t)}({\mu_{aj}}^{2} + {s_{j}}^{2})}{2{\sigma_{e}}^{2}}.
\ln \frac{\alpha_{j}}{1-\alpha_{j}}&=ln\frac{w_j}{1-w_j}+\frac{\mu_{aj}^2}{2\sigma_j^2}+\sum_{t=1}^{n}\frac{\lambda_t^n}{2\sigma_e^2}\B_{ajZ(t)}^{\prime}\B_{ajZ(t)}(\mu_{aj}^2-s_j^2+vs_j^2).
\end{align}
%Since quantifying the uncertainty of normal pattern $f_0$ is not the key here for online detection problem, we simplified the problem by estimating the point estimation of 
%with the updated $\mu_j$,$\s^2_j$ and $\alpha_j$, 
The deviation details are in Appendix A. Thus we set the posterior distribution of $\btheta_{a}$ as $\tilde{p}(\btheta_{a}|\br)=q(\btheta_{a}|\br)$ and  $\tilde{p}(\br)=q(\br)$.

Based on $\tilde{p}(\btheta_{a}|\br)$ and $\tilde{p}(\br)$, we further update the posterior distribution of $\btheta_n$. Since we assume $\btheta_{n}$ is identically and independently distributed for different $n$, the likelihood function $p((\X_{n}-\B_a\btheta_a)|\btheta_n)$ is only related to the current observation $\X_{n}$. Assume its prior also follows a Gaussian distribution $p_0(\btheta_n) \sim N(\mathbf{0},\bSigma_{b})$ with $\bSigma_{b}=\sigma_b^2\mathbf{I}$. According to Bayesian updating rule, the posterior distribution of $\btheta_n$ still follows a multivariate Gaussian distribution as $\tilde{p}(\btheta_n)\sim N(\tilde{\btheta}_n,\tilde{\bSigma}_b)$ with
\begin{align}
\label{eq:update theta_n}
\tilde{\btheta}_{n}=& (\B_{bZ(n)}^{\prime}\bSigma_{e}^{-1}\B_{bZ(n)}+\bSigma_{b}^{-1})^{-1}\B_{bZ(n)}^{\prime}\bSigma_{e}^{-1}(\X_{Z(n)}-\B_{aZ(n)}\tilde{\bmu}_{a}),\\
\label{eq:update sigma_b}
\tilde{\bSigma}_{b}=&(\B_{bZ(n)}^{\prime}\bSigma_{e}^{-1}\B_{bZ(n)}+\bSigma_{b}^{-1})^{-1}.
\end{align}
Here $\balpha=[\alpha_{1},\ldots,\alpha_{k_a}]^{\prime}$ and $\tilde{\bmu}_{a}=\bmu_{a}\circ\balpha$.

By iteratively estimating $\{\bmu_{a},\s^{2},\balpha\}$ and $\{\tilde{\btheta}_{n},\tilde{\bSigma}_{b}\}$ %$\{\tilde{\btheta}_{n},\tilde{\bSigma}_{b}\}$
until convergence, we can get $\tilde{p}(\btheta_{a},\br)$ and $\tilde{p}(\btheta_{n})$. The details of the estimation procedure are shown in Algorithm \ref{alg:VB}. 
\begin{algorithm}[h]
	%\vspace{-10pt}
	\caption{Variational Bayesian for $\btheta_{a}$ and $\btheta_{n}$}
	\label{alg:VB}
	\KwIn{Data $\X_{t}, t=1,\ldots,n$, $\btheta_{t},t=1,\ldots,n-1$}
	Initialize 
	$\tilde{\btheta}_{n}= (\B_{bZ(n)}^{\prime}\bSigma_{e}^{-1}\B_{bZ(n)}+\bSigma_{b}^{-1})^{-1}\B_{bZ(n)}^{\prime}\bSigma_{e}^{-1}\X_{Z(n)}$\\
	$\tilde{\bSigma}_{b} = (\B_{bZ(n)}^{\prime}\bSigma_{e}^{-1}\B_{bZ(n)}+\bSigma_{b}^{-1})^{-1}$ \\
	\Repeat{Converge}{
		\For{$j=1,\ldots,k_{a}$}{
			Update $\mu_{aj},s_{j}^{2}$, and $\alpha_{j}$ via (\ref{eq:update_a}), (\ref{eq:update_s2}) and (\ref{eq:update_alpha}). \\
		}
		Update $\tilde{\btheta}_{n}$ and $\tilde{\bSigma}_{b}$ via (\ref{eq:update theta_n}) and (\ref{eq:update sigma_b}). 
	}
	\Return{$\bmu_{a},\s^{2},\balpha,\tilde{\btheta}_{n}, \tilde{\bSigma}_{b}$ }
\end{algorithm}
\vspace{-10pt}
%----------------------------------------
\subsection{Bayesian Hypothesis Testing}
\label{subsec:detection}
As an alternative to classic hypothesis testing to provide evidence to support a model over another \citep{kass1995bayes}, Bayes factor uses the likelihood ratio to quantify the evidence for hypothesis $H_1$ relative to hypothesis $H_0$, i.e., $BF=\frac{\int L_1(\phi_1)\pi_1(\phi_1)d\phi_1}{\int L_0(\phi_0)\pi_0(\phi_0)d\phi_0}$, where $\phi_1=\lbrace \btheta_a,\btheta_n \rbrace$, $\phi_0=\lbrace \btheta_n \rbrace$, $\pi_i, i=1,0$ are the prior distributions and $L_j, j=1,0$ are the likelihood functions of the observations $\X_{Z(n)}$ under $H_1$ and $H_0$. It has been extensively used in model selection \citep{morey2011bayes,wasserman2000bayesian}. %\citep{morey2011bayes} presented a Bayes factor 1-sample test to conduct hypothesis test. In addition, Bayes factor can be used to conduct model selection, or conduct model averaging which estimates some quantities under each candidate model and averages the estimates according to the likelihood function of each model, and for more details see \citep{wasserman2000bayesian}. 
However, one limitation of BF is its sensitivity to variations in the prior, which may result in Lindley paradox in hypothesis testing \citep{aitkin1991posterior}. Later, as an possible solution, Posterior BF is proposed by \cite{aitkin1991posterior}. It is defined as $PBF= \frac{p(\X_{Z(n)}|H_1)}{p(\X_{Z(n)}|H_0)}=\frac{\int L_1(\phi_1)\pi_1(\phi_1|\X_{Z(n)})d\phi_1}{\int L_0(\phi_0)\pi_0(\phi_0|\X_{Z(n)})d\phi_0}$, where $\pi_i(\phi_{i}|\X_{Z(n)}), i=1,0$ are the posterior distributions under $H_{1}$ and $H_{0}$. PBF reflects the analyst's belief about the relative weighting of two competing hypotheses.
%thus a natural quantity for decision making. 

Here we construct the posterior Bayes factor based on the posterior of $\{\bmu_{a},\s^{2},\balpha,\tilde{\btheta}_{n},\tilde{\bSigma}_{b}\}$ as detection statistic to decide whether $\btheta_{a}=\mathbf{0}$. By averaging over the uncertainty of parameters, we can compute the marginal probability of the data under the two competing hypotheses, i.e.,
\begin{align}
\label{eq:original numerator}
p(\X_{Z(n)}|H_1) = &\sum_{\br \in \mathscr{R}}\tilde{p}(\br)\iint  \tilde{p}(\btheta_a|\br)\tilde{p}(\btheta_n|\btheta_a)p(\X_{Z(n)}|\btheta_a,\btheta_n)d\btheta_n d\btheta_a,\\
\label{eq:original denominator}
p(\X_{Z(n)}|H_0) = &\int \tilde{p}(\btheta_n|H_0)p(\X_{Z(n)}|\btheta_{n})d\btheta_{n}.
\end{align}
where $\mathscr{R}$ is the set of all possible values of $\br$. 
% Due to the complex form of spike-slab model, the integrals of (\ref{eq:original numerator}) and (\ref{eq:original denominator}) are intractable, especially in the high-dimensional setting. 
Then we will derive analytical forms for (\ref{eq:original numerator}) and (\ref{eq:original denominator}) in the following proposition.
\begin{proposition}
	The marginal likelihood can be estimated by integrating out the posterior distributions of the internal parameters $\{\btheta_{a},\btheta_{n}\}$ as 
	%\begin{align}
	%\label{eq:denominator}
	%\log \Big( p(\X_{Z(n)}|H_0)\Big)= &-\frac{1}{2} %\Big(\tilde{\btheta}_n^{\prime}\tilde{\bSigma}_b^{-1}\tilde{\btheta}_n+\X_{Z(n)}^{\prime}\bSigma_0\X_{Z(%n)} \\\nonumber
	%& -\mathbf{G}\mathbf{H}^{-1}\mathbf{G}^{\prime} \Big)
	%\end{align}
	\begin{align}
	\label{eq:denominator}
	&p(\X_{Z(n)}|H_0)=C_1\exp \Big(-\frac{1}{2} \big(\tilde{\btheta}_n^{[0]\prime}\tilde{\bSigma}_b^{-1}\tilde{\btheta}_n^{[0]}+\X_{Z(n)}^{\prime}\bSigma_0^{-1}\X_{Z(n)}-\mathbf{G}^{[0]}\mathbf{H}^{-1}\mathbf{G}^{[0]\prime} \big) \Big),
	\end{align}
	and 
	\begin{align}
	\label{eq:numerator}
	&p(\X_{Z(n)}|H_1)=C_2\sum_{\br \in \mathscr{R}}p(\br|H_1) \exp \Big (-\frac{1}{2}\big(\bmu_r'\mathbf{K}^{-1}\bmu_r+\X_{Z(n)}^{\prime}\bSigma_0^{-1}\X_{Z(n)}+\tilde{\btheta}_n^{[1]\prime}\tilde{\bSigma}_b^{-1}\tilde{\btheta}_n^{[1]}-\mathbf{D}\mathbf{A}^{-1}\mathbf{D}^{\prime}\\ \nonumber
	&-(\mathbf{G}^{[1]}-\mathbf{D}\mathbf{A}^{-1}\mathbf{C}^{\prime})(\mathbf{H}-\mathbf{C}\mathbf{A}^{-1}\mathbf{C}^{\prime})^{-1}(\mathbf{G}^{[1]}-\mathbf{D}\mathbf{A}^{-1}\mathbf{C}^{\prime})^{\prime}\big)\Big).
	\end{align}
	Here $\tilde{\btheta}_n^{[0]}$ equals (\ref{eq:update theta_n}) with $\tilde{\bmu}_a=0$ under $H_0$, while $\tilde{\btheta}_n^{[1]}$ equals (\ref{eq:update theta_n}) with $\tilde{\bmu}_a=\tilde{\bmu}_a$ under $H_1$. $C_1=1/\sqrt{(2\pi)^{m}|\tilde{\bSigma}_b||\bSigma_e||\mathbf{H}|}, C_2=1/\sqrt{(2\pi)^{m}|K||\tilde{\bSigma}_b||\bSigma_e||\mathbf{A}||\mathbf{H}-\mathbf{C}\mathbf{A}^{-1}\mathbf{C}^{\prime}|}$ are the constants. Some notations are defined as $\mathbf{A}=\B_{aZ(n)}^{\prime}\bSigma_e^{-1}\B_{aZ(n)}+\mathbf{K}^{-1}$, $\mathbf{D}=\X_{Z(n)}^{\prime}\bSigma_e^{-1}\B_{aZ(n)}+\bmu_r'\mathbf{K}^{-1}$, $\mathbf{C}=\B_{bZ(n)}^{\prime}\bSigma_e^{-1}\B_{aZ(n)}$, \\
	$\mathbf{H}=\B_{bZ(n)}^{\prime}\bSigma_e^{-1}\B_{bZ(n)}+\tilde{\bSigma}_b^{-1}$,$\mathbf{G}^{[0]}=\X_{Z(n)}^{\prime}\bSigma_e^{-1}\B_{bZ(n)}+\tilde{\btheta}_n^{[0]\prime}\tilde{\bSigma}_b^{-1}$,\\
	$\mathbf{G}^{[1]}=\X_{Z(n)}^{\prime}\bSigma_e^{-1}\B_{bZ(n)}+\tilde{\btheta}_n^{[1]\prime}\tilde{\bSigma}_b^{-1}$.\\
More derivation details are given in Appendix B.
\end{proposition}
By plugging (\ref{eq:denominator}) and (\ref{eq:numerator}) into $PBF_n$, we get the posterior Bayes factor, i.e., 
\begin{align}
\label{eq:bf}
&PBF_{n} =C_3\sum_{\br \in \mathscr{R}}p(\br|H_1) \exp \big( -\frac{1}{2}(\bmu_r^{\prime}\mathbf{K}^{-1}\bmu_r+\mathbf{G}^{[0]}\mathbf{H}^{-1}\mathbf{G}^{[0]\prime}-\mathbf{D}\mathbf{A}^{-1}\mathbf{D}^{\prime}-(\mathbf{G}^{[1]}-\mathbf{D}\mathbf{A}^{-1}\mathbf{C}^{\prime})\\ \nonumber
&(\mathbf{H}-\mathbf{C}\mathbf{A}^{-1}\mathbf{C}^{\prime})^{-1}(\mathbf{G}^{[1]}-\mathbf{D}\mathbf{A}^{-1}\mathbf{C}^{\prime})^{\prime}+\tilde{\btheta}_n^{[1]\prime}\tilde{\bSigma}_b^{-1}\tilde{\btheta}_n^{[1]}-\tilde{\btheta}_n^{[0]\prime}\tilde{\bSigma}_b^{-1}\tilde{\btheta}_n^{[0]})\big),
\end{align}
where $C_3=\sqrt{|\mathbf{H}|/|\mathbf{K}||\mathbf{A}||\mathbf{H}-\mathbf{C}\mathbf{A}^{-1}\mathbf{C}^{\prime}|}$. 
%Note that the form of $PBF_n$ is so difficult to compute that we can't treat it as the detection statistic directly. So we should simplify it and retain the most important terms in $PBF_n$. The simplification details are shown in Appendix B.
By dropping out some constants or extra small terms in (\ref{eq:bf}), it can be simplified. Furthermore, considering that under $H_{0}$, the terms in each $\exp(\cdot)$ of (\ref{eq:bf}) would be close to zero, we conduct Taylor expansion for further computation simplification. 
\begin{proposition}
After dropping out constants and extra small terms, we define the first order Taylor expansion of (\ref{eq:bf}) as the final detection statistic $\Lambda_n$:
\begin{align}
\label{eq:final detection}
&\Lambda_n \equiv 2\tilde{\bmu}_a^{\prime}\B_{aZ(n)}^{\prime}(\mathbf{I}-\hat{\mathbf{H}})(\X_{Z(n)}-\B_{bZ(n)}\tilde{\btheta}_n^{[1]})-\bmu_a^{\prime}(\B_{aZ(n)}^{\prime}\B_{aZ(n)}\circ\bar{\mathbf{A}})\bmu_a+\tilde{\bmu}_a^{\prime}\B_{aZ(n)}^{\prime}\hat{\mathbf{H}}\B_{aZ(n)}\tilde{\bmu}_a,
\end{align}
where $\hat{\mathbf{H}}=\B_{bZ(n)}(\B_{bZ(n)}^{\prime}\B_{bZ(n)})^{-1}\B_{bZ(n)}^{\prime}$ and $\bar{\bA}$ has diagonal items $\bar{A}_{ii}=\alpha_{i}, i = 1,\ldots, k_a$, and other items $\bar{A}_{ij}=\alpha_{i}\alpha_{j}, \forall i, j = 1,\ldots, k_a, i\neq j$. More derivation details are given in Appendix B.
\end{proposition}
For (\ref{eq:final detection}), we can set a detection threshold $h$ according to a pre-specific confidence level (false alarm rate), and define that if $\Lambda_n>h$, the test statistic triggers an abnormal alarm. Otherwise, decide next $Z(n+1)$ and wait for $\X_{n+1}$.
\vspace{-0.4em}
\subsection{Thompson Sampling for Sensor Selection}
Now we talk about how to select $Z(n+1)$. In (\ref{eq:final detection}), denote $\X_{1Z(n)}=\X_{Z(n)}-\B_{bZ(n)}\tilde{\btheta}_{n}^{[1]}$. It indicates the estimated abnormal data for the selected dimensions $Z(n)$.  When an anomaly occurs with abnormal dimension set $Z^*$, the more overlap between $Z(n+1)$ and $Z^*$, the more abnormal information $X_{1Z(n)}$ would take, and the larger value of $\Lambda_{n+1}$ is expected to be. Consequently, at the present time point, we aim to select a subset $Z(n+1)$ which can maximize the expectation of $\Lambda_{n+1}$. This is similar to the CMAB problem \citep{chen2013combinatorial}, where $Z(n+1)$ is the super arm and $\Lambda_{n+1}$ can be regarded as the reward function in our scenario.
%Recall that $\X_{Z(t)}=\B_{bZ(t)}\btheta_{t}+\B_{aZ(t)}\btheta_a+\mathbf{E}_{Z(t)}$.
%Note that the decision statistic $\Lambda_{n+1}$ contains the item $\X_{Z(n+1)}-\B_{bZ(n+1)}\tilde{\btheta}_{n+1}^{[1]}$, indicating that the value of the decision statistic is decided by the randomness of $\btheta_a$ and $\mathbf{E}$, but regardless of $\btheta_n$.
% Since $\btheta_{t}$ and $\mathbf{E}_{Z(t)}$ are $i.i.d.$ for $t=1,2,\ldots$, they can be seen as contexts in contextual multi-armed bandit problem. The expected reward $E(\Lambda_{n+1})$ is the function of sampling set $Z(n+1)$, the contexts of arms $\{\B_{bi}\btheta_{n+1}\}_{i=[p]}$, $\{\mathbf{E}_{i}\}_{i=[p]}$, $\B_{ai}$ and the expected scores $\{\B_{ai}\btheta_a\}_{i=[p]}$ associated with each arm $i$, where $[p]=\{1,...,p\}$.
%In a system, there are $p$ arms, each can generate reward from an unknown distribution with an unknown mean. Same as multi-armed bandit(MAB) problem, the objective of combinatoral multi-armed bandit problem is to repeatedly play these arms in multiple rounds so that the total expected reward is as close to the total reward of playing the optimal arm as possible. Unlike MAB, the arms are not played one by one, but to play a subset of arms in a round which form certain combinatorial structures. If we treat each subset of arms as a super arm, these super arms are dependent with each other because of the combinatorial structures. After playing a super arm, the outcomes of related arms are observed, we use these information to select the super arm for the next round. 
Following the Bayesian estimation framework, we propose to construct the sequential decision of $Z(n+1)$ based on Thompson sampling, which has been shown to perform competitively to the state of the arts in a variety of bandit and adaptive sampling problems \citep{agrawal2012analysis,agrawal2013thompson}. Under the framework of Thompson sampling, based on the current inference of $\X_{1}$ so far, the probability of $\hat{Z}$ to maximize $\Lambda_{n+1}$ is
\begin{align}
\label{eq:thompson}
\int \mathbb{I}\Big(\hat{Z}=\arg\max_{Z}(\Lambda_{n+1}\vert Z,\X_{1})\Big)f(\X_{1})d\X_{1},
\end{align}
where $\mathbb{I}$ is the indicator function and $f(\X_{1})$ is the posterior distribution of $\X_{1}$, which can be calculated from $\tilde{p}(\btheta_a,\mathbf{r})$. 
The core of the Thompson sampling is to sample a $\hat{\X}_{1}$ from $f(\X_{1})$ instead of computing the integral in (\ref{eq:thompson}). This can be achieved by sampling a $\hat{\btheta}_a$ from $\tilde{p}(\btheta_a,\mathbf{r})$, sampling a $\hat{\mathbf{E}}$ from its distribution $N(\mathbf{0},\bSigma_e)$ and getting $\hat{\X}_{1}=\B_a\hat{\btheta}_a+\hat{\mathbf{E}}$.
% This can be achieved by sampling a $\hat{\btheta}_n$ from $\tilde{p}(\btheta_n)$, sampling a $\hat{\btheta}_a$ from $\tilde{p}(\btheta_a,\mathbf{r})$, sampling a $\hat{\bE}$ from its distribution $N(\mathbf{0},\bSigma_e)$ and getting $\hat{\X}=\B_b\hat{\btheta}_n+\B_a\hat{\btheta}_a+\hat{\bE}$. 
Then select $Z(n+1)$ according to
%$\hat{\btheta}_{a}$ from the posterior distribution of $\tilde{p}(\btheta_{a},\br)$ and sample $\hat{\mathbf{E}}$ from $N(\mathbf{0},\bSigma_0)$. Since the anomaly term and the noise term are independent, get a sample of $\hat{\X}_2$ as $\B_{a}\hat{\btheta}_{a}+\hat{\mathbf{E}}$. Replace the corresponding $\X_{n2}$ by $\hat{\X}_2$  in (\ref{eq:simplified reward function}), as $\Lambda_{n2}(\hat{\X}_2)$. Since $\hat{\X}_{2}$ is sampled from its distribution, it can explore the values of $\btheta_a$ where the anomaly may occur. Thus the objective is find the optimal subset $Z^*$, so it also do exploitation,
\begin{align}
\label{eq:thompson sampling}
&Z(n+1)=\max_{Z}\Lambda_{n+1}(\hat{\X}_1)=2\tilde{\bmu}_a^{\prime}\B_{aZ}^{\prime}(\mathbf{I}-\hat{\mathbf{H}})\B_{aZ}\hat{\X}_1-\bmu_a^{\prime}(\B_{aZ}^{\prime}\B_{aZ}\circ\bar{\mathbf{A}})\bmu_a+\tilde{\bmu}_a^{\prime}\B_{aZ}^{\prime}\hat{\mathbf{H}}\B_{aZ}\tilde{\bmu}_a.
\end{align}
Hereafter we define the strategy of (\ref{eq:thompson sampling}) as the oracle sampling procedure. The random sampling procedure encourages the exploration, and the maximization of $\Lambda_{n+1}$ encourages the exploitation. Therefore, the proposed approach achieves good balance between exploration and exploitation. Furthermore, it has a good property that the regret between $Z^\star$ and $Z(n+1)$ converges to zero as $n$ goes on. Though the strategy of (\ref{eq:thompson sampling}) is desirable and can achieve good performance (as shown in Section \ref{sec:generalized experiments}), one limitation is that the sampling process requires large computation complexity, since $\Lambda_{n+1}(\hat{\X}_1)$ is a nonlinear function of $Z(n+1)$ and all the $p\choose{m}$ subsets need to be evaluated for best selection. This is a common problem for many CMAB strategies \citep{chen2016combinatorial}, where they usually assume an oracle computer center can evaluate all the combinations. However, enumerating all possible combinations of the arms is intractable especially when the number of arms is large. Therefore, we would like to reduce the complexity of the CMAB problem through the following proposition.
\begin{proposition}
	\label{proposition:ortho}
Suppose any column of $\B_a$ or $\B_b$, denoted as $\bb$, satisfies $\lVert \bb ^2 \rVert_\infty \leq \frac{c}{p}\lVert \bb \rVert_2^2$, where $c\in \mathbb{R}_+$ is a constant satisfying $1 \leq c\leq p$. Let $\epsilon,\delta \in (0,1]$ be two small values, and let $\frac{c^2}{2\epsilon^2}log(\frac{(k_a+k_b)^2}{\delta}) \leq m \leq \frac{2a_p^2p^2\epsilon^2}{c^2log(\frac{(k_a+k_b)^2}{\delta}}$ be an integer, where $a_p$ is the smallest probability that a variable can be sampled. For all possible $m$-dimensional subsets $Z$ in $\mathscr{Z}=\left\{Z_k,k=1,...,M\right\}$ where $M =$$p\choose{m}$, with probability at least $1-2\delta$, for any column $\bb_{ai}$ of $\B_a$ and any column $\bb_{bj}$ of $\B_b$, we have
	\begin{align*}
	-a_p\epsilon \leq \bb_{aiZ}^{\prime}\bb_{bjZ}\leq a_1\epsilon,
%	-a_p\epsilon \leq \mathbf{B}_{aiZ}^{\prime}\mathbf{B}_{ajZ}\leq a_1\epsilon
	\end{align*}
	where $0 \leq a_p \leq a_1 \leq 1$ are small constants. This indicates $\mathbf{B}_{bZ}^{\prime}\mathbf{B}_{aZ}=\mathbf{0}$ are approximately valid. 
	%imilarly, $\mathbf{B}_{aZ}^{\prime}\mathbf{B}_{aZ}=\bI$ can be proved.
	The verification details are shown in Appendix C.
\end{proposition}
%This assumption of the orthogonality on the normal bases is valid since we can always perform orthogonalization on a set of bases\cite{yan2017anomaly}. 
%Finally, the assumption of the orthogonality constraint on $m$-sparse subsets are inspired by the restricted isometry property, which implies the matrices are nearly orthogonal when acting on the sparse vectors. 
Under Proposition \ref{proposition:ortho}, the sampling procedure of (\ref{eq:thompson sampling}) can be further simplified as:
%(However, it can also be relaxed and the proposed framework would still work), i.e., 
\begin{align}
\label{eq:final expected reward}
Z(n+1)=\max_{Z}\Lambda_{n+1}(\hat{\X}_1)&=2\tilde{\bmu}_{a}^{\prime}\B_{aZ}^{\prime}\B_{aZ}\hat{\X}_1-\bmu_{a}^{\prime}(\B_{aZ}^{\prime}\B_{aZ}\circ\bar{\mathbf{A}})\bmu_{a}\\ \nonumber
&=\sum_{i\in Z}\Big(2\hat{\X}_1^{\prime}\B_{ai}^{\prime}\B_{ai}\tilde{\bmu}_a-\bmu_a^{\prime}(\B_{ai}^{\prime}\B_{ai}\circ \bar{\mathbf{A}})\bmu_a\Big),
\end{align}
where $\B_{ai}$ is the $i^{th}$ row of $\B_{a}$. This indicates that we only need to rank 
\begin{align}
\Lambda_{(n+1)i}=2\hat{\X}_1^{\prime}\B_{ai}^{\prime}\B_{ai}\tilde{\bmu}_a-\bmu_a^{\prime}(\B_{ai}^{\prime}\B_{ai}\circ \bar{\mathbf{A}})\bmu_a, i=1,...,p
\end{align}
from the largest to the smallest and select the top $m$ items, instead of enumerating all the possible sets of $Z$. Hereafter we denote (\ref{eq:final expected reward}) as the simplified sampling procedure. Consequently, the complexity of sampling process drops dramatically and allows us to handle very high dimensional data. The simplified procedure of Thompson sampling is shown in Algorithm \ref{alg:thom} with the total computation for one time point as $O(p\log(p))$.
\begin{algorithm}
	\caption{Simplified Thompson Sampling Procedure}
	\label{alg:thom}
	\KwIn{$\tilde{p}(\btheta_a,\br)$, $\tilde{\bmu}_{a}$, $\bmu_{a}$, $\B_{aZ(n)}$, $\bar{\mathbf{A}},\bSigma_e$ estimated upon to the current time point $n$}
	\KwOut{$Z(n+1)$}
	Sample $\hat{\btheta}_a \sim \tilde{p}(\btheta_a,\br)$, sample $\hat{\mathbf{E}}$ from $N(\mathbf{0},\bSigma_e)$ and get $\hat{\X}_1=\B_a\hat{\btheta}_a+\hat{\mathbf{E}}$\\ 
	Compute $\Lambda_{(n+1)i}=2\hat{\X}_1^{\prime}\B_{ai}^{\prime}\B_{ai}\tilde{\bmu}_a-\bmu_a^{\prime}(\B_{ai}^{\prime}\B_{ai}\circ \bar{\mathbf{A}})\bmu_a$, $i=1,2,...p$\\
	Rank the $\Lambda_{(n+1)i}$, for $i=1,2,...p$ from the largest to the smallest, and select the top $m$ items as $Z(n+1)$\\
%	Apply $Z(n+1)$, observe $\X_{Z(n+1)}$ and update $\tilde{p}(\btheta_a,\br)$.
\end{algorithm}

\subsection{Theoretical Properties}
Now we discuss some theoretical properties of the simplified Thompson sampling procedure under asymptotic conditions. These properties are built upon the asymptotic properties of variation Bayesian inferences in \cite{wang2019frequentist}. We first consider the specific cases when there is no background information $\B_{b}\btheta_{n}$ in the data stream. In this case, we do not need to estimate $\btheta_{n}$ at all, and can directly set (\ref{eq:update theta_n}) and (\ref{eq:update sigma_b}) to be zero. 
\begin{theorem}
\label{theorem:in control sampling} For a system without background information, i.e., $\B_{b}\btheta_{n}=\mathbf{0}$,  when there is no change in the system, as $n \rightarrow \infty$, we have  $q_j(\theta_{aj}) \xrightarrow{d} \delta_0$, $\forall j=1...k_a$, where $\delta_0$ is a point mass at $0$. This means $\mu_{aj}\rightarrow 0$ and $s_j^2\rightarrow 0$. Consequently we have  $E(\Lambda_{(n+1)i})\rightarrow 0$ and $Var(\Lambda_{(n+1)i})\rightarrow 0$, $\forall i=1...p$. 
% $\Lambda_{i}=2(\hat{\btheta}_a^{\prime}\B_{a}^{\prime})_{i}\B_{ai}\tilde{\bmu}_a-\B_{ai}(\bmu_a\bmu_a'\circ \bar{\mathbf{A}})\B_{ai}^{\prime}$ follows Gaussian distribution. If satisfy the following conditions, when the process is normal, $\Lambda_i$ are independently and identically distributed for all $i$. $\B_b^{\prime}\B_a=\mathbf{0}$, $\B_a^{\prime}\B_a=\I$, and under Proposition \ref{proposition:ortho}, $\B_{bZ(t)}^{\prime}\B_{aZ(t)}=\mathbf{0}$ and $\B_{aZ(t)}^{\prime}\B_{aZ(t)} =\I$ are approximately valid. For all $j$, the values of entries of $\B_{aj}$ are the same except positions. For all $i$, the values of entries of $\B_{ai}$ are the same except positions. 
\end{theorem}
More details are shown in Appendix D. Theorem \ref{theorem:in control sampling} indicates that when the system has no change, under the limit conditions, Algorithm \ref{alg:thom} can select $Z(n+1)$ from all variables randomly. 

\begin{theorem}
\label{theorem:out of control sampling}
For a system without background information, i.e., $\B_{b}\btheta_{n}=\mathbf{0}$, when the system has change, assume the change relates to certain bases $\calA\subset \{1,\ldots,k_{a}\}$ with change magnitude $\phi_{l}, l\in \calA$. As $n \rightarrow \infty$, 
$q_l(\theta_{al}) \xrightarrow{d} \delta_{\phi_l}$ for $l\in \calA$ where $\delta_{\phi_l}$ is a point mass at $\phi_l$, and $q_j(\theta_{aj}) \xrightarrow{d} \delta_0, \forall j \notin \calA$. This means $\mu_{al}\rightarrow \phi_l$, $\alpha_{l}\rightarrow 1$ and $s_l^2\rightarrow 0, \forall l \in \calA$. As to other bases, $\mu_{aj}\rightarrow 0$ and $s_j^2\rightarrow 0$, $\forall j \notin \calA$. Consequently, $E(\Lambda_{(n+1)i})\rightarrow \sum_{l \in \calA}B_{ail}^2\phi_l^2+2\sum_{l_1,l_2\in \calA, l_1 \neq l_2}B_{ail_1}B_{ail_2}\phi_{l_1}\phi_{l_2}$ and $Var(\Lambda_{(n+1)i})\rightarrow 0$, $\forall i=1...p$.
% However, after an anomaly with magnitude $\delta$ related to the $l^{\mathrm{th}}$ base $\B_{al}$ appears in the system, we obtain
% 	\begin{align}
% 	\label{eq:oc mean of lambda}
% E(\Lambda_{i})&=O(B_{ail}^2\mu_{al}^2\alpha_{l})\\
% \label{eq:oc variance of lambda}
% Var(\Lambda_{i})&=O\Big(B_{ail}^2s_l^2(\sum_{j=1}^{k_a} B_{aij}\mu_{aj}\alpha_j)^2\Big)
% 	\end{align}
\end{theorem}
% From (\ref{eq:oc mean of lambda}), we can see that the variables which have larger entries in $\B_{al}$ will have larger $\Lambda_{ni}$, so we sample the most related variables to the $l^{\mathrm{th}}$ base preferentially and consistently. Therefore $\B_a$ should be designed according to the interested patterns. If one of these patterns occurs in the system, then its most related anomalous variables can be selected and this pattern can be detected out. The detailed proof of this theorem is shown in Appendix D. For sensitivity analysis, if $m/p$ increases, $\alpha_l$ will increase to be closer to $1$ and $\mu_{al}$ will increase to be closer to $\delta$. So $E(\Lambda_{i})=O(B_{ail}^2\mu_{al}^2\alpha_{l})$ will be larger and can trigger an alarm earlier while keeping the in-control $ARL_0$ invariant. If $\delta$ increases, $\mu_{al}$ will increase. So $E(\Lambda_{i})=O(B_{ail}^2\mu_{al}^2\alpha_{l})$ will also be larger and can trigger an alarm earlier while keeping the in-control $ARL_0$ invariant.
More details are shown in Appendix D. Theorem \ref{theorem:out of control sampling} indicates that in abnormal condition, we prefer to choose the variables mostly influenced by the abnormal patterns. 

As to general cases with background information, we have the following Corollary \ref{col:IC_OC}, where the condition $m \to \infty$ is to guarantee $\btheta_{n}$ can be estimated accurately. Then the properties of estimated $\btheta_{a}$ together with $\Lambda_{(n+1)i}$ in Theorem \ref{theorem:in control sampling} and Theorem \ref{theorem:out of control sampling} can be guaranteed. 
\begin{corollary}
\label{col:IC_OC}
For a general system with background information $\B_{b}\btheta_{n}$, under $p \to \infty$, $m \to \infty$ (but $m/p$ can be bounded or go to infinity), the results of Theorem \ref{theorem:in control sampling} and Theorem \ref{theorem:out of control sampling} hold. 
\end{corollary}

Hereafter, we shorten our proposed Composite Decomposition based Spike and Slab Detection scheme using oracle sampling procedure as CDSSD(O) and using simplified sampling procedure as CDSSD. The full detection scheme is shown in Algorithm \ref{alg:Detection}. %Here the control limit $h$ is set according to numerical simulations such that 
%The complexity of Algorithm \ref{alg:Detection} is $O(k_a\times niter)+O(p)+O(2^{k_a})$, where $niter$ is the iteration times when updating posterior, so the total computational cost is almost linear in the dimension of observations. Therefore, the proposed method is scalable in online detection of high-dimensional stream data. 
\begin{remark}
	It is to be noted that in some real applications where the prior information of $\B_{a}$ is unknown and not inferable, $\B_{a}$ can be simply set to be identity matrix $\B_{a}=\bI \in \calR^{p \times p}$. In this case it aims to detect sparse changes on the original $p$ dimensions directly.  
\end{remark}
%-----------------------------------
\begin{algorithm}
	\caption{Composite Decomposition based Change Detection}
	\label{alg:Detection}
	\KwIn{Data streams $\X_{n}, n=1,\ldots$}
	Set the initial sampling set $Z(1)$ by randomly selecting $m$ variables out of the $p$ variables.\\ 
	\For{$n= 1,\ldots$}{
		Collect the data $\X_{Z(n)}$
		Update posterior distributions via Algorithm \ref{alg:VB}. \\
		Calculate the detection statistic $\Lambda_{n}$ via Eq (\ref{eq:final detection}).  \\
		\eIf{$\Lambda_{n}>h$}{
			Trigger a change alarm}
		{Decide the next sampling set $Z(n+1)$ via Algorithm \ref{alg:thom}.
		}
	}
\end{algorithm}
\section{Numerical Studies}
\label{sec:simulation}
In this section, we conduct extensive experiments on both synthetic and real-world data sets to evaluate the performance of our proposed CDSSD. We also compare it with the following existing baselines:\\
\textbf{TRAS}: top-r adaptive sampling detection algorithm in \citet{liu2015adaptive}.\\
\textbf{NAS}: nonparametric anti-rank adaptive sampling algorithm in \citet{xian2017nonparametric}.\\
\textbf{CMAB(s)}: simplified combinatorial multi-armed bandit adaptive sampling strategy in \citet{zhang2019partially}.\\
\textbf{SASAM}:	spatial-adaptive sampling and monitoring procedure in \citet{wang2018spatial}.\\
\textbf{CDSSD(I)}: a variant strategy of CDSSD, which sets the anomaly bases $\B_{a}$ to be the identity matrix. This represents cases when $\B_{a}$ is neither known nor inferable.\\
\textbf{ORACLE}: the proposed detection scheme yet assuming all the $p$ variables are fully observable at each time point and no adaptive sampling is required. It is used as a performance upper bound of our detection scheme.
\subsection{Efficiency comparison between CDSSD and CDSSD(O)}
\label{sec:generalized experiments}
We first evaluate the performance difference between CDSSD and CDSSD(O) and show that the simplification of CDSSD has little influence to the detection power, and yet can save a lot of computation. 
% Also, to show the influence of Assumption \ref{proposition:ortho} on detection power, we also conduct the generalized Thompson sampling based on (\ref{eq:thompson sampling}) without Assumption \ref{assumption:ortho}, which is shorten as CDSSD(G). 
At the same time, we also compare them with the combinatorial multi-armed bandit (CMAB) method proposed by \cite{zhang2019partially}. Because of high computation complexity of CDSSD(O) and CMAB(s), we only use low dimensions of $p=15,m=5,8,11$, and compare ADDs of these three strategies.  

In our simulation, we assume $\B_{b}\in R^{15\times 3}$ are the three lowest frequency Fourier bases and $\B_{a}\in R^{15 \times 10}$ are $10$ four-order B-spline bases with $14$ equally spaced knots. When the data is normal, $\btheta_{t}\sim N(\mathbf{0},\bSigma_{b})$ with $\bSigma_{b}=\sigma_{b}^{2}\bI$ and $\sigma_{b}=0.3$. $\mathbf{E}_{t}\sim N(\mathbf{0},\bSigma_{e})$ with $\bSigma_{e}=\sigma_{e}^{2} \bI$ and $\sigma_{e} =0.05$. When a change occurs after the change point $\tau=50$, we assume only the $j^{\mathrm{th}}$ column of the $10$ B-spline bases has nonzero coefficient $\theta_{aj}=\phi$, where $\phi$ is the change magnitude. For each simulation replication, we set $j$ by randomly drawing a number from $1$ to $10$, and generate random samples of $\X_{t}=\B_b\btheta_t+\B_a\btheta_a+\bE_t$ for a total time length $T=2000$ from the above experimental settings. For CMAB(s), the parameters are set according to the algorithm in \cite{zhang2019partially}. For CDSSD, we set $\lambda=0.1$, $w_j=0.1$, $\sigma_j=3$ for $j=1,...,k_a$, and $v=10^{-7}$. For all the methods, we tune their detection thresholds to ensure that their $ARL_0$ is exactly $200$ such that their detection performance under change cases can be fairly compared. Then we record the first time point that each algorithm triggers a change alarm as its corresponding detection delay. We calculate ADD using $1000$ replications, as the performance evaluation criterion of different methods. 

The results are shown in Table \ref{tab:s1}. We can see that both CDSSD(O) and CDSSD strongly outperform CMAB(s) in all $\phi$'s magnitudes and $m/p$ settings, demonstrating the superiority of our proposed methods. The difference between CDSSD(O) and CDSSD is not significant. Only when $m/p$ is quite small and the magnitude of defect $\phi$ is quite small, the gap between CDSSD and CDSSD(O) is obvious. In other cases, CDSSD performs almost as well as CDSSD(O). This indicates that in most scenarios, CDSSD can be served as a substitute of CDSSD(O). Therefore in the following experiments, we only compare CDSSD with other state-of-the-art methods for performance evaluation for computation reduction. 
\begin{table}[htbp]
	\vspace{-0.1in}
	\caption{ADDs (and Standard Deviation of Detection Delays, i.e., STDs in the abbreviation) for Experiments of $\X_{t}\in \calR^{15\times 1}$. }
	\label{tab:s1}
	%\vskip -0.2in
	\vspace{-0.1in}
	\begin{center}
		\begin{sc}
			\scalebox{0.8}{
				\hspace{-0.5in}
				\begin{tabular}{p{0.4cm}cccccccccc}
					\toprule
					& \multicolumn{3}{c}{$p=15,m=5$} & \multicolumn{3}{c}{$p=15,m=8$} &\multicolumn{3}{c}{$p=15,m=11$}\\
					$\phi$ & \footnotesize{CDSSD} & \footnotesize{CDSSD(O)}& \footnotesize{CMAB(s)} & \footnotesize{CDSSD} & \footnotesize{CDSSD(O)}& \footnotesize{CMAB(s)} & \footnotesize{CDSSD} & \footnotesize{CDSSD(O)} & \footnotesize{CMAB(s)} \\
					\cmidrule(lr){2-4} \cmidrule(lr){5-7} \cmidrule(lr){8-10}
					0.0& 200(251)& 200(227)&200(187)&200(267)& 200(273)&200(187)& 200(295)& 200(272)&200(193)\\
					0.1& 30.3(37.4)& \textbf{16.4}(16.8)&59.6(51.6)& 13.2(15.2)& \textbf{9.00}(9.29)&21.8(13.6)& 7.06(7.73)& \textbf{5.94}(5.89)&12.9(7.22)\\
					0.2& 8.16(7.59)& \textbf{5.76}(4.58)&24.3(16.6)& 3.91(3.39)& \textbf{3.17}(2.24)&8.41(4.10)& 2.48(2.08)& \textbf{2.07}(1.36)&5.02(1.84)\\
					0.3& 4.58(4.05)& \textbf{3.73}(2.68)&14.7(9.23)& 2.20(1.62)& \textbf{1.96}(1.16)&5.42(2.51)& 1.56(1.03)& \textbf{1.39}(0.68)&3.29(1.19)\\
					0.4& 3.49(2.73)& \textbf{2.79}(2.11)&10.5(6.62)& 1.77(1.29)& \textbf{1.59}(0.93)&3.88(1.69)& 1.28(0.70)& \textbf{1.19}(0.48)&2.47(0.76)\\
					0.5& 2.87(2.35)& \textbf{2.53}(2.07)&8.22(5.36)& 1.60(1.25)& \textbf{1.41}(0.76)&3.01(1.37)& 1.25(0.68)& \textbf{1.12}(0.41)&1.96(0.62)\\
					0.6& 2.50(1.93)& \textbf{2.17}(1.67)&6.37(4.24)& 1.47(1.04)& \textbf{1.33}(0.67)&2.54(1.24)& 1.21(0.69)& \textbf{1.10}(0.41)&1.96(0.62)\\
					0.7& 2.23(1.67)& \textbf{2.06}(1.46)&5.38(3.68)& 1.39(0.94)& \textbf{1.31}(0.72)&2.15(1.12)& 1.19(0.67)& \textbf{1.07}(0.28)&1.44(0.55)\\
					0.8& 2.11(1.57)& \textbf{1.93}(1.50)&4.71(3.15)& 1.39(0.96)& \textbf{1.23}(0.60)&1.92(0.93)& 1.15(0.58)& \textbf{1.07}(0.33)&1.23(0.43)\\
					0.9& 2.09(1.72)& \textbf{1.80}(1.27)&4.26(2.89)& 1.32(0.82& \textbf{1.22}(0.60)&1.79(0.93)& 1.15(0.63)& \textbf{1.07}(0.31)&1.14(0.38)\\
					1.0& 1.96(1.50)& \textbf{1.79}(1.26)&3.57(2.38)& 1.30(0.86)& \textbf{1.16}(0.50)&1.61(0.76)& 1.11(0.46)& \textbf{1.06}(0.29)&1.07(0.27)\\
					\bottomrule
			\end{tabular}}
		\end{sc}
	\end{center}
	\vspace{-0.1in}
\end{table}
\subsection{One-Dimensional(1D) Experiments}
\label{sec:numerical1}
In this section, we consider higher dimensional cases with $\X_{t}\in \calR^{30\times 1}$. We assume $\B_{b}\in R^{30\times 2}$ are the two lowest frequency Fourier bases and $\B_{a}\in R^{30 \times 17}$ are $17$ four-order B-spline bases with $21$ equally spaced knots. All the other experimental parameters including $\btheta_{b}$, $\btheta_{a}$, $\mathbf{E}_{t}$ are generated in the same way as Section \ref{sec:generalized experiments}. As to other baseline methods, for TRAS, we set its parameters $r=m$, $\mu_{min}=0.05$ and $\Delta=0.0001$ according to recommendation of \cite{liu2015adaptive}. For NAS, we set $k=0.05,\Delta=0.07$ following the algorithm in \cite{xian2017nonparametric}. For SASAM, the parameters are selected to be $\theta_1=0.1$, $\theta_2=0.7$, $h=1$ and $\mu_{min}=0.1$ according to the recommendations of \cite{wang2018spatial}.  

\begin{figure*}[htbp]
	\begin{subfigure}[b]{0.32\textwidth} 
		\includegraphics[width=5.5cm,height=3.5cm]{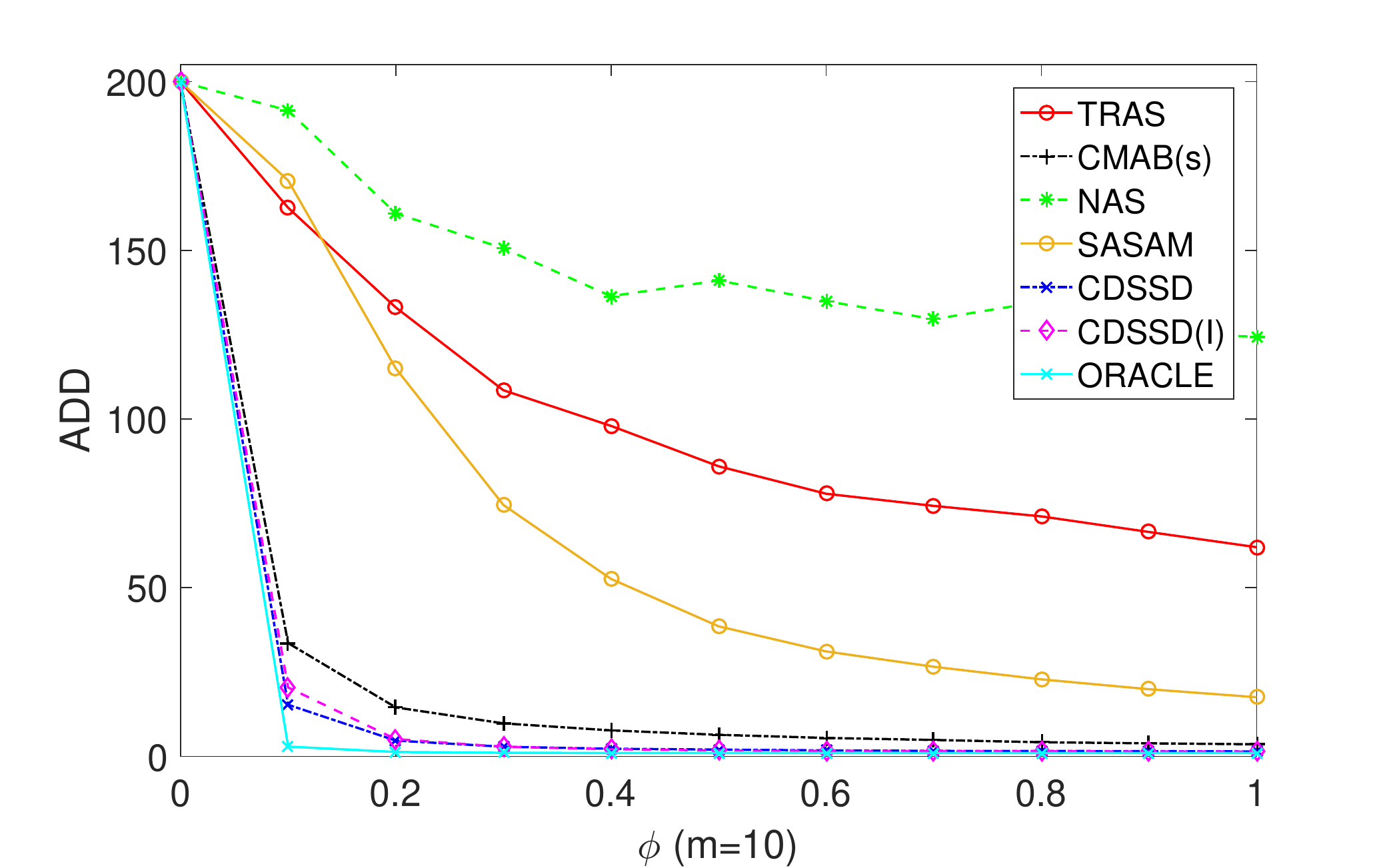}
		\caption{}
		\label{fig:fig1}
	\end{subfigure}
	\begin{subfigure}[b]{0.32\textwidth} 
		\includegraphics[width=5.5cm,height=3.5cm]{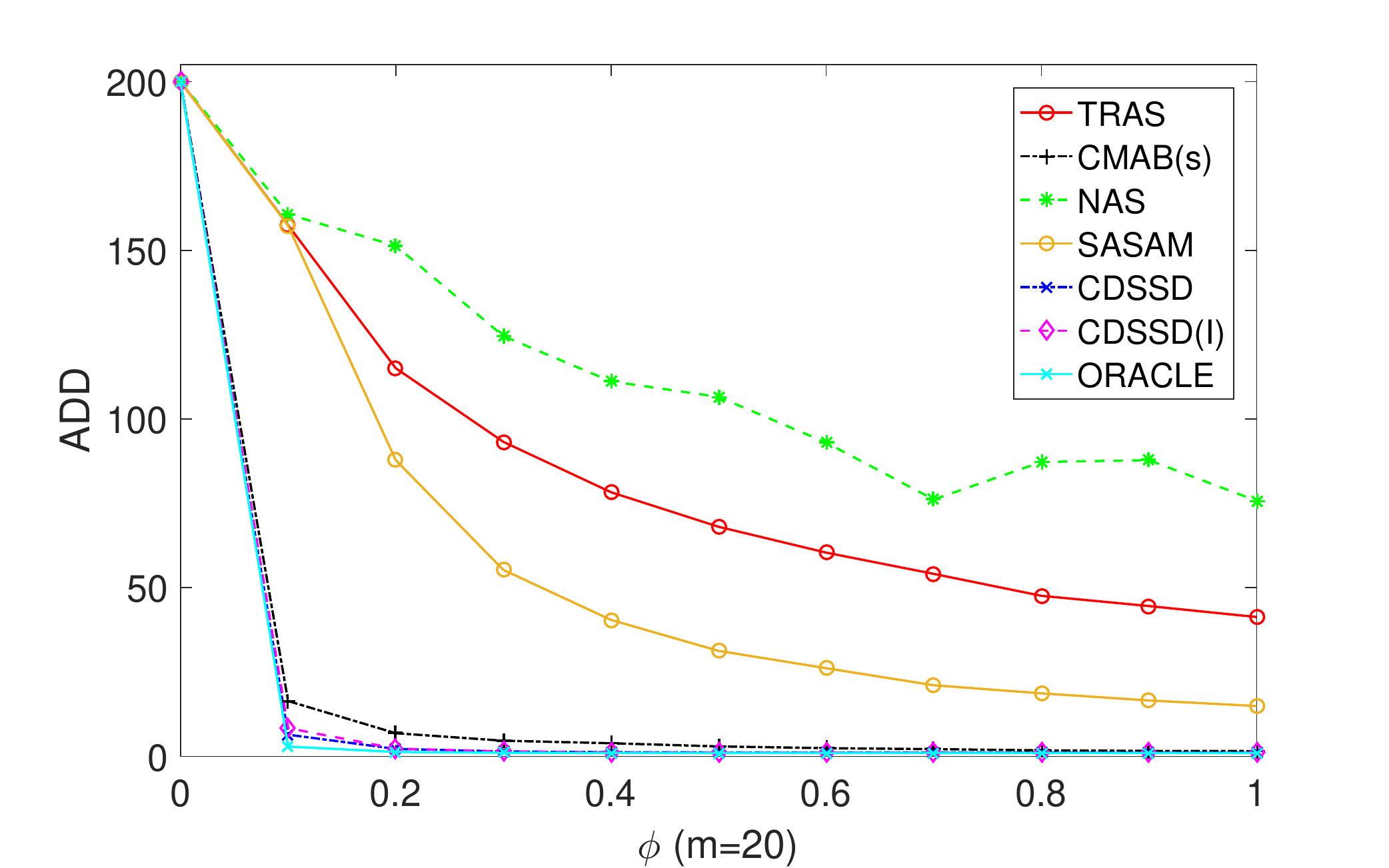}
		\caption{}
		\label{fig:fig2} 
	\end{subfigure}
	\begin{subfigure}[b]{0.32\textwidth} 
		\includegraphics[width=5.5cm,height=3.5cm]{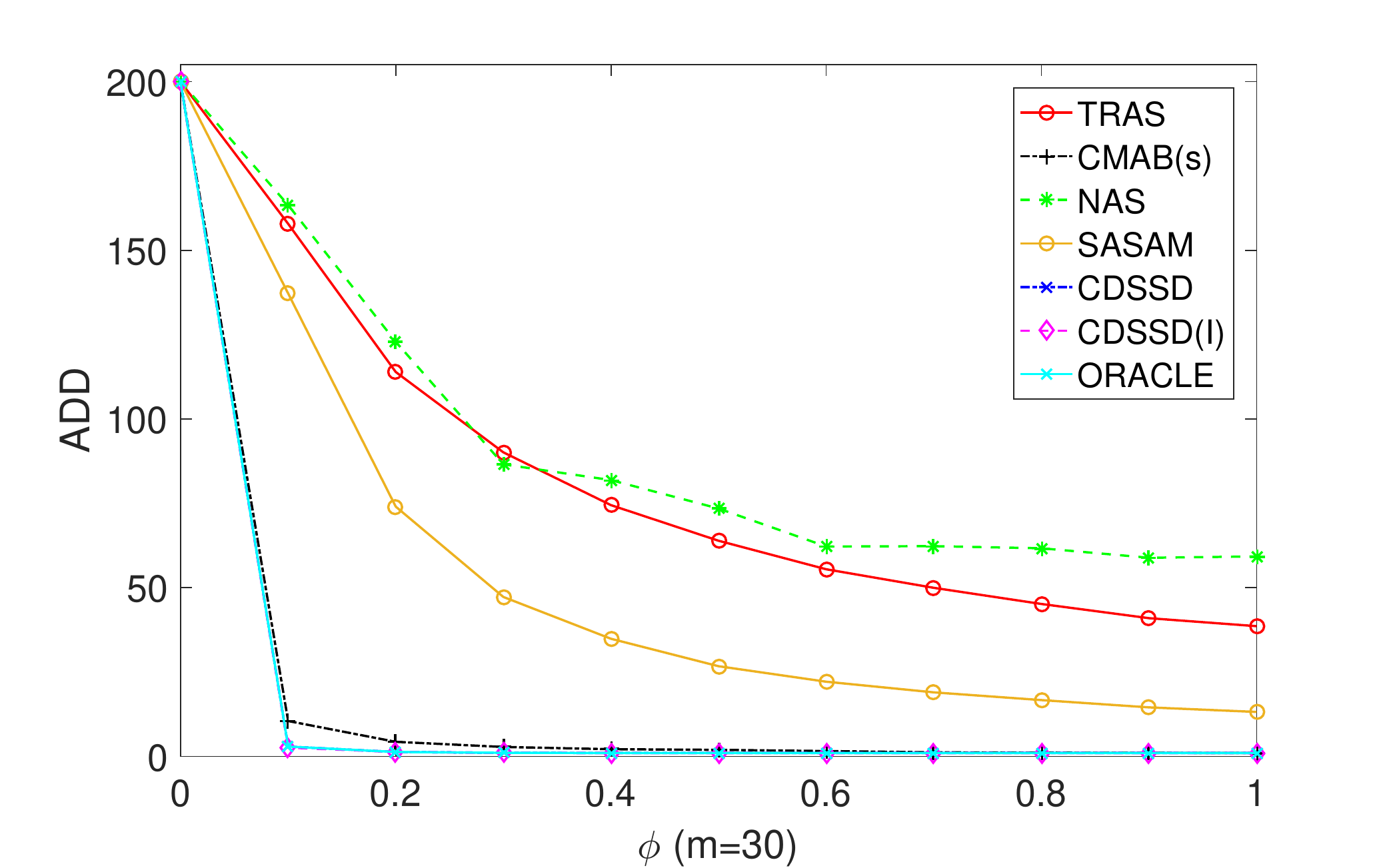}
		\caption{}
		\label{fig:fig3} 
	\end{subfigure}
	\caption{ADDs for 1D experiments with (a) $m=10$, (b) $m=20$, (c) $m=30$}
	\label{} 
\end{figure*}

The detection results for $\phi$ ranging from $0$ to $1$ with $m=10$, $20$ and $30$ are shown in Figures \ref{fig:fig1}, \ref{fig:fig2} and \ref{fig:fig3} respectively. The detailed values together with their standard deviations are shown in Appendix E. Clearly, except ORACLE, which is infeasible in practice, CDSSD has the smallest ADD generally, followed by CDSSD(I), demonstrating their detection power of our proposed detection framework. In particular, for small $\phi$, CDSSD performs better than CDSSD(I), while for large $\phi$, is slightly inferior to CDSSD(I). This is because for small $\phi$, the change pattern as a whole contributes to the detection. While when $\phi$ is larger, it is certain individual dimensions distinctly influenced by anomaly patterns that contribute to the detection statistic mostly.
% And since in B-spline bases, variables have nonzero entries on several adjacent bases, when $\phi$ is large, CDSSD may cause false abnormal identification of the adjacent bases to the actual abnormal base, which may lead to a little inefficient sampling process. 
Consequently CDSSD(I) with identity $\B_{a}$ can also have satisfactory detection performance. CMAB(s) performs a little inferiorly compared with CDSSD or CDSSD(I), followed by SASAM. As to CUSUM and NAS, their performances are not very satisfactory, since they do not consider either correlations of different variables or change sparsity.  
\subsection{Extension to Two-Dimensional (2D) Experiments}
\label{sec:numerical2}
\begin{figure*}[htbp]
	\begin{subfigure}[b]{0.32\textwidth} 
		\includegraphics[width=5.5cm,height=3.5cm]{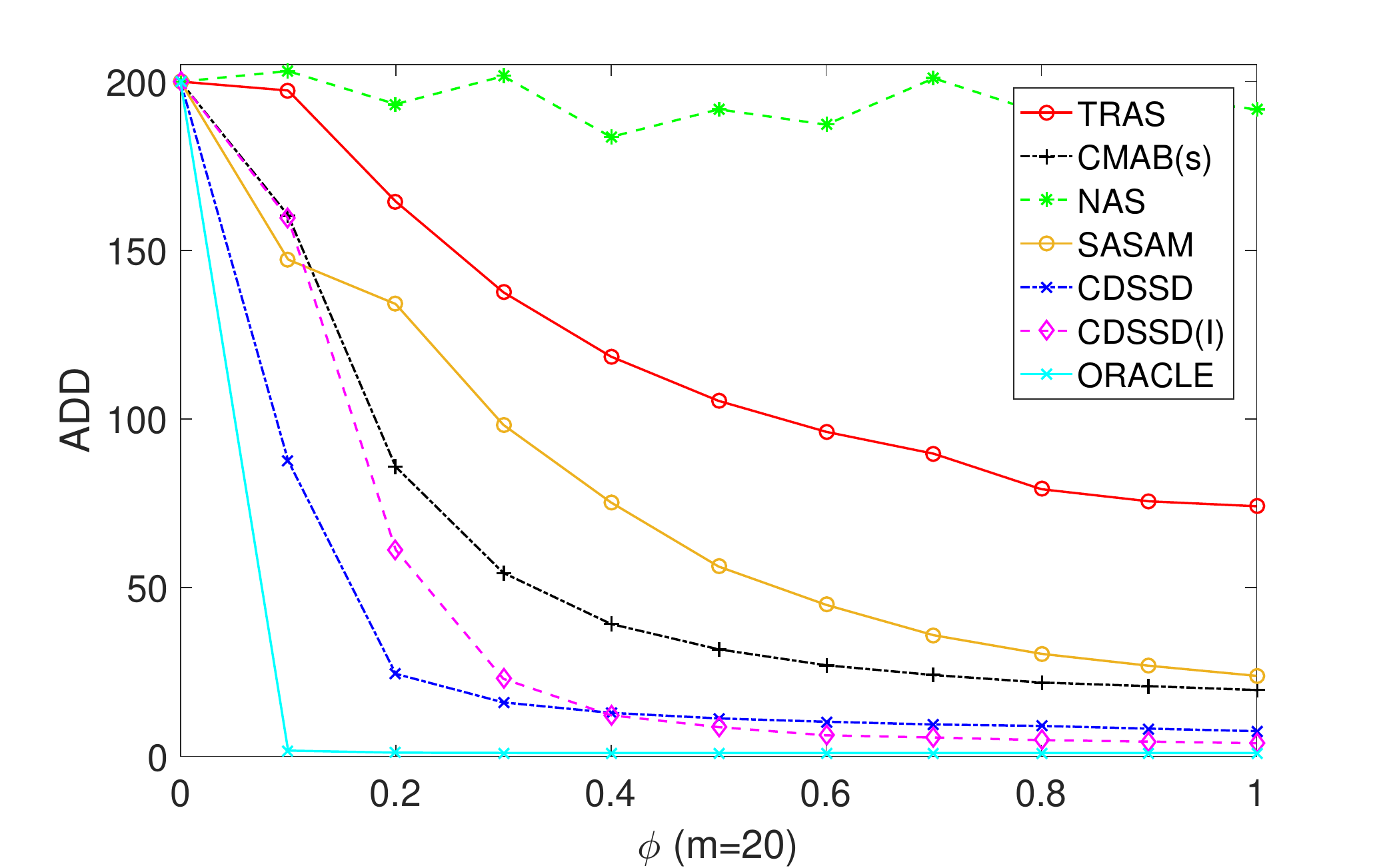}
		\caption{}
		\label{fig:fig4}
	\end{subfigure}
	\begin{subfigure}[b]{0.32\textwidth} 
		\includegraphics[width=5.5cm,height=3.5cm]{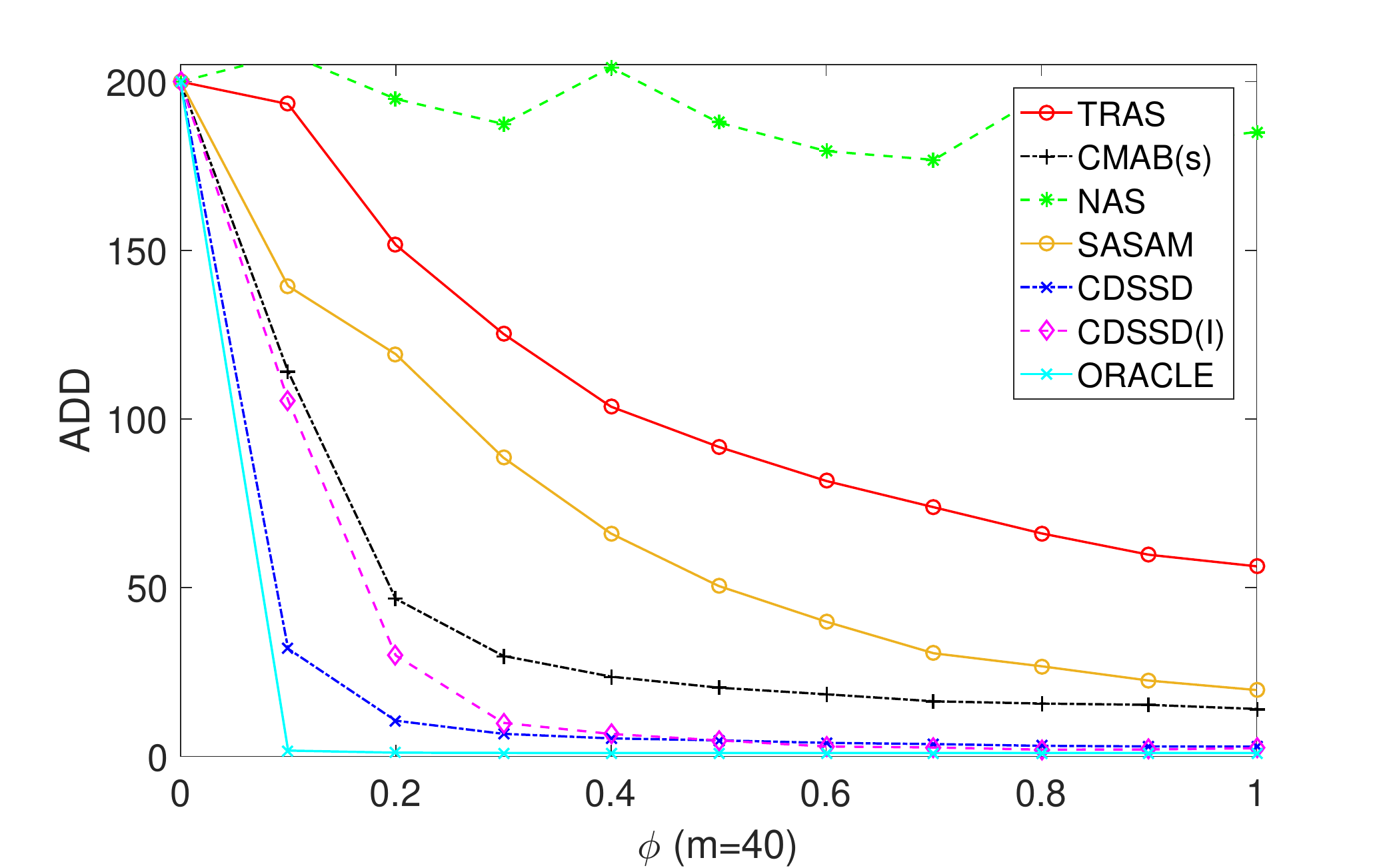}
		\caption{}
		\label{fig:fig5} 
	\end{subfigure}
	\begin{subfigure}[b]{0.32\textwidth} 
		\includegraphics[width=5.5cm,height=3.5cm]{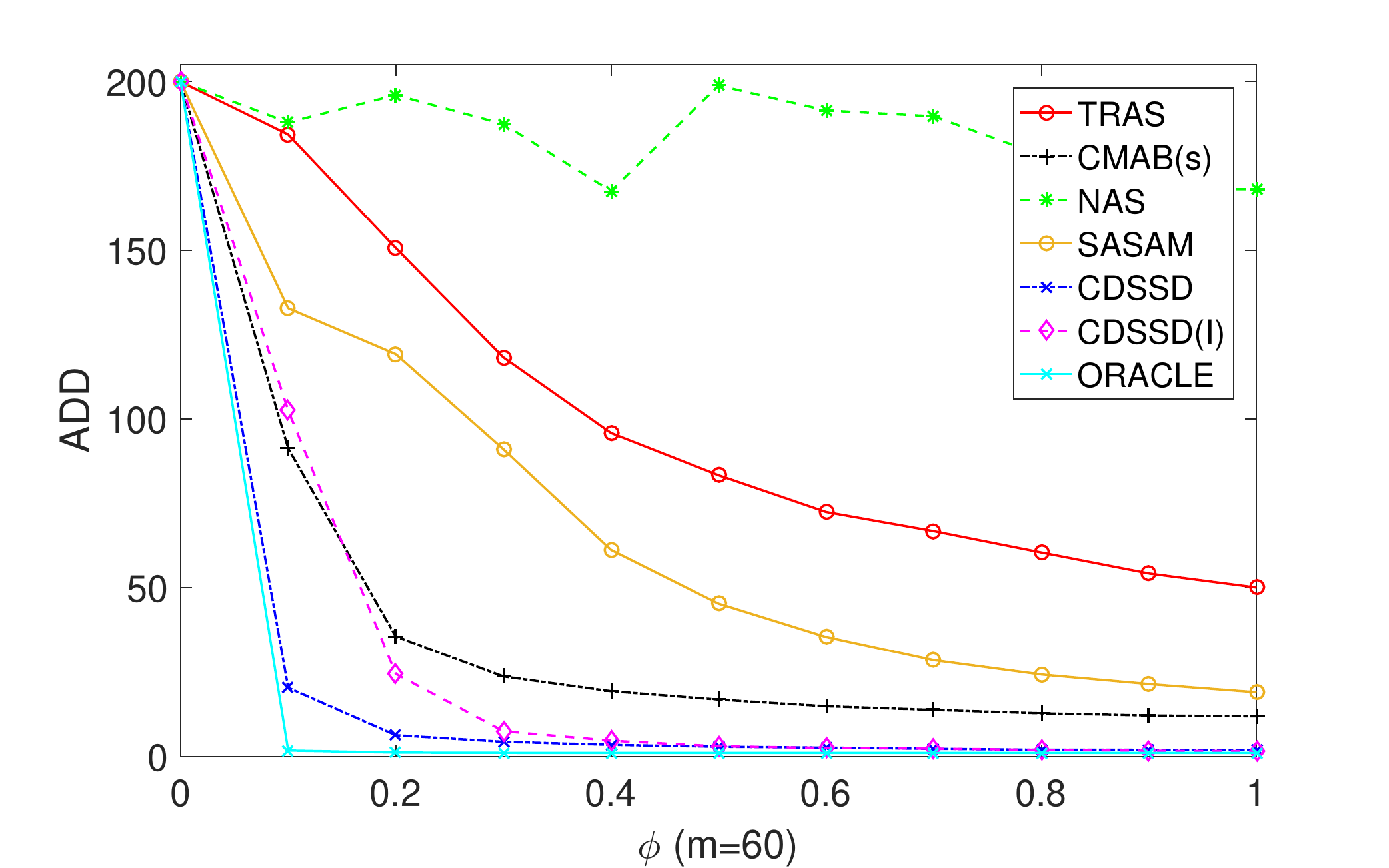}
		\caption{}
		\label{fig:fig6} 
	\end{subfigure}
	\caption{ADDs for 2D experiments with (a) $m=20$, (b) $m=40$, (c) $m=60$}
	\label{} 
\end{figure*}

In this experiment, we further consider data with more complex spatial structure, i.e.,  $\X_{t}$ as an image with $20 \times 20$ pixels.  We first generate each column of $\bb_{b}\in \calR^{20 \times 2}$ from two-order B-spline bases with $4$ equally spaced knots, and set $\B_{b}=\bb_{b}\otimes\bb_{b}$, where $\otimes$ is the Kronecker tensor product. Similarly, we generate each column of $\bb_{a}\in \calR^{20 \times 9}$ from four-order B-spline bases with $13$ equally spaced knots, and set $\B_{a}=\bb_{a}\otimes \bb_{a}$. All the other experimental parameters including $\btheta_{b}$, $\btheta_{a}$, $\mathbf{E}_{t}$ are generated in the same way as Section \ref{sec:generalized experiments}. 
We also tune the change magnitude $\phi$ and evaluate the performance of different algorithms according to their ADDs.
%given that their detection threshold can ensure their ADD$_{0}$ equal to $200$.
We vectorize each $\X_{t}$ as a vector with $p = 400$ to construct the detection statistics for all the methods. The performance of different methods under $m=20$, $40$ and $60$ is shown in Figures \ref{fig:fig4}, \ref{fig:fig5} and \ref{fig:fig6}. The specific ADD values together with their standard deviations are added in Appendix E. Similar as the result in Section \ref{sec:numerical1}, CDSSD performs the best generally, but is slightly inferior to CDSSD(I) when $m/p$ is small and $\phi$ is large, due to the same reason as Section \ref{sec:numerical1}. However, its gap from ORACLE is larger than that of one dimensional case in Section \ref{sec:numerical1}. This is because the proportion of observable dimensions, i.e., $m/p$, is much smaller than that of one dimensional case. In addition, other methods, i.e., CMAB(s), SASAM, CUSUM and NAS perform worse than CDSSD and even CDSSD(I).

\subsection{Case Study on Solar Flare Detection}
%Solar flare is the sudden flash of increased brightness on the surface or the edge of the sun. It can release ultraviolet rays and X-rays carrying high energy, which may intefere the radio communication on the earth, or cause damage to the space station and the astronauts. To reduce these damage, it is important to detect and predict the occurrence of the solar flare as early as possible. In the space station, thousands of snapshots of the sun are captured by satellite every second. Such fast sampling frequency leads to the data we obtain and use to analyze every second is really a huge amount. That is quite unrealistic since the transmisstion bandwidth from space station to earth control center, or the processing speed of the computer or the memory of the computer is limited.
We apply CDSSD to a real case study, i.e., the solar flare detection with the same data set as \cite{liu2015adaptive,wang2018spatial}. 
%We propose to use CDSSD to tackle this problem. The solar flare data is public available online, and was also studied by \citep{liu2015adaptive} and \citep{wang2018spatial} in their case studies. 
The data set is in video format and contains $300$ frames of sequential images, each of which have $67744$ pixels distributed on a $232 \times 292$ grid. By treating one pixel as a dimension, the total vectorized data have dimension $67744$. %And online observations form a data stream to be processed in real time. 
The solar flare appears at time $t=187 \sim 202$ in this data set. %Applying the idea of data decompostion, we should construct normal bases and abnormal bases, and then use CDSSD to estimate and detect anomaly. 
We use the first 100 frames as training data for parameter estimation. In particular, we conduct principal component analysis for $t=1,\ldots, 10$, and extract the first $20$ principal components as the  dictionary of normal bases, i.e., $\B_{b}\in \calR^{67,744 \times 20}$, %according to which we construct the normal bases, $\B_{b} \in \calR^{67,744 \times 20}$. 
The extracted PCA scores represent $\btheta_{t}$, and can be used to further compute the prior covariance matrix of $\btheta_{t}$, i.e., $\bSigma_b$, and the standard deviation of noise, i.e., $\sigma_e=0.0319$. As for the anomaly bases, it is desirable to construct $\B_{a}$ according to the size and shape of possible patterns, which can be obtained from historical abnormal data when solar flares occur. Consider that the pattern of solar flare approximates to small circle piles, and forms many free shapes by these circle piles. According to this prior information, we generate $\B_{a}\in \calR^{67,744\times 256}$ from three-order B-spline bases with $19$ equally spaced knots.
%$\bb_{a}\in \calR^{292 \times 16}$ from three-order B-spline bases with $19$ equally spaced knots, and get $\B^{temp}_{a}=\bb_{a}\otimes \bb_{a}\in \calR^{85,264\times 256}$. Finally, we delete the excess rows of $\B^{temp}_{a}$, whose number is corresponding to the location of the excess points converting from $292\times292$ image to $232\times292$ image, and thus get our desired anomaly bases, $\B_{a}\in \calR^{67,744\times 256}$.

%To compare the effectiveness and efficiency of our CDSSD with the state-of-the-art methods, we need to tune their in-control ARL$_0$ to a prescribed value such that their false alarm under the in-control case is the same. In this case, 
%Considering that we only have $300$ observations, to estimate the detection threshold $h$, we need large enough in-control samples. \citep{qiu2008distribution}and \citep{qiu2011nonparametric} applied Monte Carlo simulation and used a bisection search algorithm to calculate it. \citep{liu2015adaptive} used bootstrap technique to generate a dataset of $5000$ samples by randomly drawing the data with replacement from the first $100$ in control samples. Here, we also use bootstrap method to generate $3000$ samples from the first $100$ in control samples before the occurrence of the anomaly and the $203^{rd} \sim 215^{th}$ samples after the occurrence of the anomaly.
We assume that only $400$ out of $67744$ pixels are available in our case, while \cite{liu2015adaptive} assumed $2000$ out of $67744$ observable and \cite{wang2018spatial} set $500$ out of $67744$ pixels observable . 
To show the detection efficacy of CDSSD, we compare it with the other methods in this literature. We set $ARL_{0}=1100$ according to \cite{wang2018spatial} for all the methods. According to the requirements in \citep{liu2015adaptive}, the parameters of TRAS are selected to be $\mu_{min}=2.1$ and $\Delta=5*10^{-6}$. The parameters of SASAM are set to be $\theta_1=0.1$, $\theta_2=0.4$, $h=5$ and $\mu_{min}=2$ according to recommendations in \citep{wang2018spatial}. As for NAS, the parameters are selected $\Delta=1.47*10^{-5}$, $k=0.1$, $\lambda=1.6*10^{-3}$ and $\lambda_0=0.0103$ according to \citep{xian2017nonparametric}. For our method CDSSD, we set the parameters $\sigma_j=10$, $w_j=0.1$ for $j=1,2,...,k_a$ and $\lambda=0.1$ according to the requirements of the algorithm. Here we don't compare with CMAB(s) in \citep{zhang2019partially} since in CMAB(s), we need to construct the covariance matrix of size of $67744 \times 67744$. That requires more than $32$ GB memory of computer, which is really time-consuming to implement and thus inefficient for online detection.

\begin{figure}[htbp]
	\centering
	\begin{subfigure}[b]{0.242\textwidth} 
		\centering
		\includegraphics[width=3.5cm,height=2.3cm]{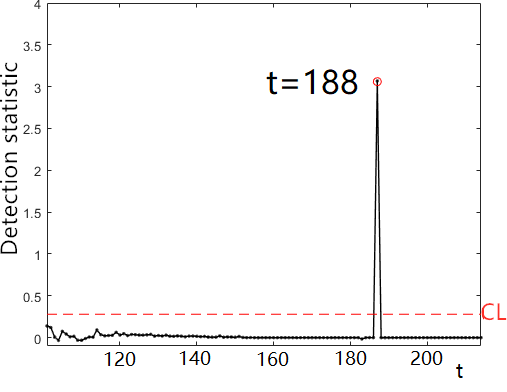}
		\caption{}
		\label{} 
	\end{subfigure}
	\begin{subfigure}[b]{0.242\textwidth} 
		\centering
		\includegraphics[width=3.5cm,height=2.3cm]{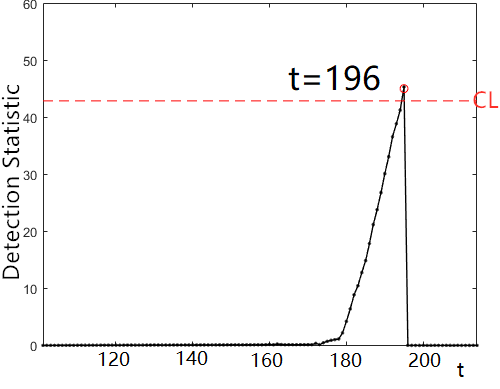}
		\caption{}
		\label{} 
	\end{subfigure}
%	\hfill
	\begin{subfigure}[b]{0.242\textwidth} 
		\centering
		\includegraphics[width=3.5cm,height=2.3cm]{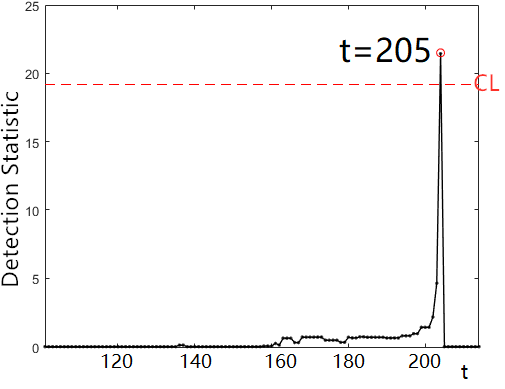}
		\caption{}
		\label{}
	\end{subfigure}
	\begin{subfigure}[b]{0.242\textwidth} 
		\centering
		\includegraphics[width=3.5cm,height=2.3cm]{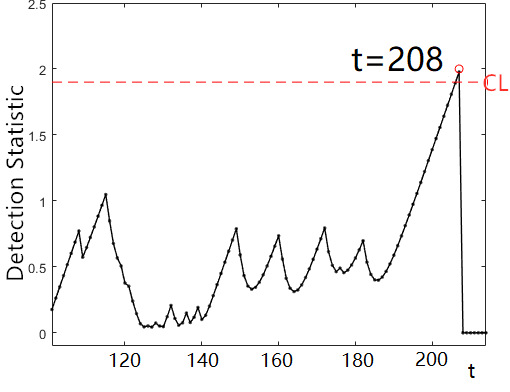}
		\caption{}
		\label{}
	\end{subfigure}
	\caption{Detection statistics for solar flare case (a) $DD_{CDSSD}=2$, (b) $DD_{TRAS}=10$, (c) $DD_{SASAM}=19$ and (d) $DD_{NAS}=22$.}
	\label{fig:test} 
\end{figure}

\begin{figure}[htbp]
	\centering
	\begin{subfigure}[b]{0.45\textwidth} 
		\centering
		\includegraphics[width=6cm,height=4.8cm]{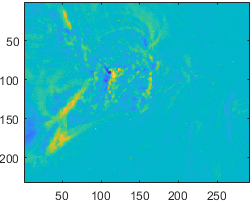}
		\caption{}
		\label{} 
	\end{subfigure}
	\begin{subfigure}[b]{0.45\textwidth} 
		\centering
		\includegraphics[width=6cm,height=4.8cm]{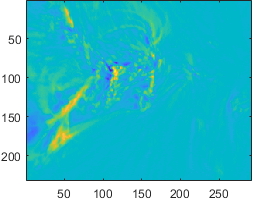}
		\caption{}
		\label{} 
	\end{subfigure}
%	\hfill

	\begin{subfigure}[b]{0.45\textwidth} 
		\centering
		\includegraphics[width=6cm,height=4.8cm]{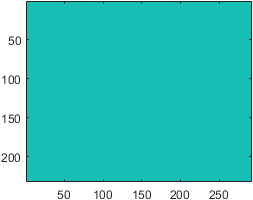}
		\caption{}
		\label{}
	\end{subfigure}
	\begin{subfigure}[b]{0.45\textwidth} 
		\centering
		\includegraphics[width=6cm,height=4.8cm]{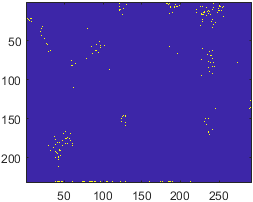}
		\caption{}
		\label{}
	\end{subfigure}
	\caption{Detection results of solar flare case at $t=180$.(a) Original data (b) Fitted normal data (c) Fitted abnormal data (d) Sampling points}
	\label{fig:t180} 
\end{figure}

\begin{figure}[htbp]
	\begin{subfigure}[b]{0.45\textwidth} 
		\centering
		\includegraphics[width=6cm,height=4.8cm]{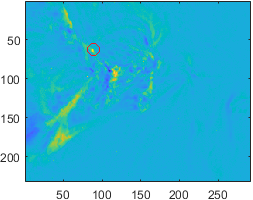}
		\caption{}
		\label{} 
	\end{subfigure}
	\begin{subfigure}[b]{0.45\textwidth} 
		\centering
		\includegraphics[width=6cm,height=4.8cm]{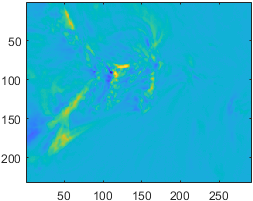}
		\caption{}
		\label{} 
	\end{subfigure}
%	\hfill

	\begin{subfigure}[b]{0.45\textwidth} 
		\centering
		\includegraphics[width=6cm,height=4.8cm]{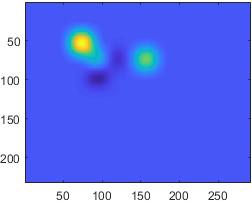}
		\caption{}
		\label{}
	\end{subfigure}
	\begin{subfigure}[b]{0.45\textwidth} 
		\centering
		\includegraphics[width=6cm,height=4.8cm]{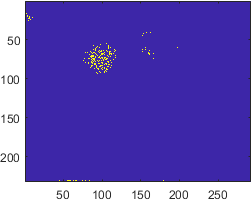}
		\caption{}
		\label{}
	\end{subfigure}
	\caption{Detection results  of solar flare case at $t=188$.(a) Original data (b) Fitted normal data (c) Fitted abnormal data (d) Sampling points}
	\label{fig:t188} 
\end{figure}

\begin{figure}[htbp]
	\begin{subfigure}[b]{0.45\textwidth} 
		\centering
		\includegraphics[width=6cm,height=4.8cm]{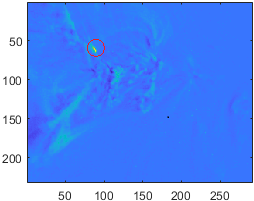}
		\caption{}
		\label{} 
	\end{subfigure}
	\begin{subfigure}[b]{0.45\textwidth} 
		\centering
		\includegraphics[width=6cm,height=4.8cm]{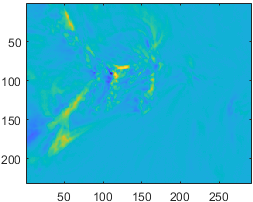}
		\caption{}
		\label{} 
	\end{subfigure}
%	\hfill

	\begin{subfigure}[b]{0.45\textwidth} 
		\centering
		\includegraphics[width=6cm,height=4.8cm]{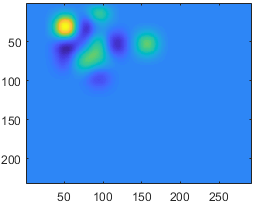}
		\caption{}
		\label{}
	\end{subfigure}
	\begin{subfigure}[b]{0.45\textwidth} 
		\centering
		\includegraphics[width=6cm,height=4.8cm]{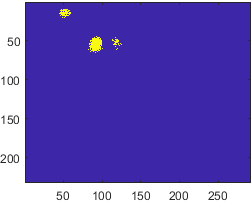}
		\caption{}
		\label{}
	\end{subfigure}
	\caption{Detection results of solar flare case at $t=192$.(a) Original data (b) Fitted normal data (c) Fitted abnormal data (d) Sampling points}
	\label{fig:t192} 
\end{figure}

The monitoring process starts from $t=101$. The DDs(Dection Delay) of the four methods are $DD_{CDSSD}=2$, $DD_{TRAS}=10$, $DD_{SASAM}=19$ and $DD_{NAS}=22$ respectively. Their detection statistics are shown in Figure \ref{fig:test}. As we can see, CDSSD has the smallest DD$=2$, outperforming other methods and achieving efficient online anomaly detection.
% \begin{table}
% 	\begin{center}
% 		\caption{ADDs of different methods on the real data set}
% 		\label{tab:table2}
% 		\begin{tabular}{cccc}
% 			\hline
% 			CDSSD &TRAS &SASAM &NAS\\
% 			\hline
% 			2 &10 &19 &22\\
% 			\hline
% 		\end{tabular}
% 	\end{center}
% \end{table}
To better illustrate the  performance of CDSSD, we visualize its detection results of three selected time points $t=180,188$ and $192$. When the anomaly has not occur at $t=180$, Figure \ref{fig:t180} (a) shows the figure of sun's surface in normal condition.
The detection result indicates that there is no fitted anomaly pattern in Figure \ref{fig:t180} (c) and the sampling points are distributed randomly in Figure \ref{fig:t180} (d). After the solar flare occurs at $t=186$ which is strengthened in red circle in Figure \ref{fig:t188} (a) and Figure \ref{fig:t192} (a), At $t=188$, CDSSD first detects the anomaly. As we can see, there appears a fitted anomaly pattern in Figure \ref{fig:t188} (c) and the sampling points concentrate at the area of the solar flare in Figure \ref{fig:t188} (d). To show this is not a short-time concentration like random sampling, we further check the detection results after triggering an alarm, e.g., at $t=192$. The detection results are the same as that of $t=188$. So we can conclude that before anomaly appears, CDSSD searches all the variables randomly and does not concentrate any set of variables. However, after anomaly appears, CDSSD can concentrate on the variables influenced by the anomaly for a period of time.  
\section{Conclusions}
This paper addresses high dimensional sequential change detection problem with partial observations. It proposes an adaptive sampling method to select a subset of variables in the system for online monitoring. 
Specifically, to deal with the correlations among variables and sparse changes in the system, we introduce the framework of sparse smooth composite decomposition, based on which we learn the value of parameters via spike and slab variational Bayesian inference. To be coherent, using the estimated parameters, we construct the posterior Bayesian factor as our detection statistic. By formulating the detection statistic as the reward function in multi-armed bandit problem, we propose a Thompson sampling strategy for sampling the most informative variables for the next time point. This sampling strategy can achieve two desirable properties, (1) randomly sampling among variables when the process is normal and (2) sampling anomalous points preferentially and consistently when change appears in the system. So it can achieve good exploration and exploitation property, which contributes greatly to the efficiency of our proposed algorithm. Finally, through synthetic and real-world data experiments, we conclude that our method performs much better than existing adaptive sampling strategies. In the area of online process monitoring, this research develops an novel adaptive sampling strategy to determine which subset of data streams should be observed when only a limited number of resources are available. The applications of our proposed method are extensive, such as syndromic surveillance in epidemiology, network traffic control and surveillance video. 
% This paper proposes a novel detection framework for POSSCD based on composite decomposition. We learn the decomposition via spike-and-slab variational Bayesian inference based on partial observations, and construct consequent detection scheme by posterior Bayes factor. By formulating the test statistic as the reward function in multi-armed bandit problem, we propose a Thompson sampling strategy for the most informative sampling for the next time point. At last, we use experiments on both synthetic and real-world data set to show the efficacy and applicability of our method.
\begin{small}
	\bibliography{adaptive_vb}
	\bibliographystyle{asa}
\end{small}

\clearpage
\section*{Appendix A: Deviation of Bayesian Inference}
\label{Appendix1}
We slightly abuse the notation by writing 
\begin{align*}
&\tilde{\X}_{Z(t)}=\X_{Z(t)}-\B_{bZ(t)}\btheta_t.
\end{align*}
The joint posterior distribution $p(\tilde{\X},\btheta_a,\mathbf{r})$ and its logarithm transformation can be expressed as
\begin{align*}
p(\btheta_a,\mathbf{r},\tilde{\X}_{Z(1)},\tilde{\X}_{Z(2)}...\tilde{\X}_{Z(n)})&=\prod_{t=1}^{n}p(\tilde{\X}_{Z(t)}|\btheta_a,\mathbf{r})^{\lambda_t^n}\prod_{j=1}^{k_a}p(\theta_{aj}|r_j)\prod_{j=1}^{k_a}p(r_j), \\
\ln p(\btheta_a,\mathbf{r},\tilde{\X}_{Z(1)},\tilde{\X}_{Z(2)}...\tilde{\X}_{Z(n)})&=\lambda_t^n\sum_{t=1}^{n}\ln p(\tilde{\X}_{Z(t)}|\btheta_a,\mathbf{r})+\sum_{j=1}^{k_a}\ln p(\theta_{aj}|r_j)+\sum_{j=1}^{k_a}\ln p(r_j).
\end{align*}
Further, the first part of $p(\tilde{\X},\btheta_a,\mathbf{r})$ can be derived as
\begin{align*}
\prod_{t=1}^{n}p(\tilde{\X}_{Z(t)}|\btheta_a,\mathbf{r})^{\lambda_t^n}&=\prod_{t=1}^{n}\Big(\frac{1}{(2\pi)^{p/2}\sigma_e}\exp\big(-\frac{(\tilde{\X}_{Z(t)}-\B_{aZ(t)}\btheta_a)^{\prime}(\tilde{\X}_{Z(t)}-\B_{aZ(t)}\btheta_a)}{2\sigma_e^2}\big)\Big)^{\lambda_t^n}, \\
\lambda_t^n\sum_{t=1}^{n}\ln p(\tilde{\X}_{Z(t)}|\btheta_a,\mathbf{r})&=\lambda_t^n\sum_{t=1}^{n}\Big((pc-\frac{1}{2}\ln\sigma_e^2)-\frac{1}{2\sigma_e^2}\big(\tilde{\X}_{Z(t)}^{\prime}\tilde{\X}_{Z(t)}-2\tilde{\X}_{Z(t)}^{\prime}\sum_{j}\B_{ajZ(t)}\theta_{aj}\\
&+(\sum_{j}\B_{ajZ(t)}\theta_{aj})^{\prime}(\sum_{j}\B_{ajZ(t)}\theta_{aj})\big)\Big),
\end{align*}
where $c=-\frac{\ln(2\pi)}{2}$. To compute negative Kullback-Leibler divergence between $p(\tilde{\X},\btheta_a,\mathbf{r})$ and $q(\btheta_a,\mathbf{r})$, take its expectation under the distribution of $q(\btheta_a,\mathbf{r})$
\begin{align*}
&E_{q(\btheta_a,\mathbf{r})}\Big(\lambda_t^n\sum_{t=1}^{n}\ln p(\tilde{\X}_{Z(t)}|\btheta_a,\mathbf{r})\Big)=(pc-\frac{1}{2}\ln\sigma_e^2)-\sum_{t=1}^{n}\frac{\lambda_t^n}{2\sigma_e^2}\Big(\tilde{\X}_{Z(t)}^{\prime}\tilde{\X}_{Z(t)}-2\tilde{\X}_{Z(t)}^{\prime}\sum_{j}\B_{ajZ(t)}\alpha_j\mu_{aj}\\
&+2\sum_j\sum_{k\neq j}\B_{ajZ(t)}^{\prime}\B_{akZ(t)}\alpha_j\alpha_k\mu_{aj}\mu_{ak}+\sum_j\B_{ajZ(t)}^{\prime}\B_{ajZ(t)}\big((\mu_{aj}^2+s_j^2)\alpha_j+vs_j^2(1-\alpha_j)\big)\Big).
\end{align*}
Also, the second part of $p(\tilde{\X},\btheta_a,\mathbf{r})$ can be derived as
\begin{align*}
\prod_{j=1}^{k_a}p(\theta_{aj}|r_j)&=\prod_{j=1}^{k_a}\Big(\frac{1}{\sqrt{2\pi}\sigma_j}\exp(-\frac{1}{2\sigma_j^2}\theta_{aj}^2)\Big)^{I(r_j=1)}\Big(\frac{1}{\sqrt{2\pi v}\sigma_j}\exp(-\frac{1}{2v\sigma_j^2}\theta_{aj}^2)\Big)^{I(r_j=0)}, \\
\sum_{j=1}^{k_a}\ln p(\theta_{aj}|r_j)&=\sum_{j=1}^{k_a}\Big((c-\frac{1}{2}\ln\sigma_j^2-\frac{\theta_{aj}^2}{2\sigma_j^2})r_j+(c-\frac{1}{2}\ln(v\sigma_j^2)-\frac{\theta_{aj}^2}{2v\sigma_j^2})(1-r_j)\Big).
\end{align*}
Take its expectation under the distribution of $q(\btheta_a,\mathbf{r})$
\begin{align*}
&E_{q(\btheta_a,\mathbf{r})}\Big(\sum_{j=1}^{k_a}\ln p(\theta_{aj}|r_j)\Big)=\sum_{j=1}^{k_a}\Big(\alpha_j(c-\frac{1}{2}\ln\sigma_j^2-\frac{\mu_{aj}^2+s_j^2}{2\sigma_j^2})+(1-\alpha_j)(c-\frac{1}{2}\ln(v\sigma_j^2)-\frac{vs_j^2}{2v\sigma_j^2})\Big).
\end{align*}
And the third part of $p(\tilde{\X},\btheta_a,\mathbf{r})$ can be derives as
\begin{align*}
&\prod_{j=1}^{k_a}p(r_j)=\prod_{j=1}^{k_a}w_j^{r_j}(1-w_j)^{1-r_j},\\
&\sum_{j=1}^{k_a}\ln p(r_j)=\sum_{j=1}^{k_a}\Big(r_j\ln(w_j)+(1-r_j)\ln(1-w_j)\Big).
\end{align*}
Take its expectation under the distribution of $q(\btheta_a,\mathbf{r})$
\begin{align*}
E_{q(\btheta_a,\mathbf{r})}\Big(\sum_{j=1}^{k_a}\ln p(r_j)\Big)=\sum_{j=1}^{k_a}\Big(\alpha_j\ln(w_j)+(1-\alpha_j)\ln(1-w_j)\Big).
\end{align*}
To sum up, the expectation of joint posterior distribution $p(\tilde{\X},\btheta_a,\mathbf{r})$ under the distribution of $q(\btheta_a,\mathbf{r})$ is
\begin{align*}
&E_{q(\btheta_a,\mathbf{r})}\Big(\ln p(\btheta_a,\mathbf{r},\tilde{\X}_{Z(1)},\tilde{\X}_{Z(2)}...\tilde{\X}_{Z(n)})\Big)=(pc-\frac{1}{2}\ln\sigma_e^2)-\sum_{t=1}^{n}\frac{\lambda_t^n}{2\sigma_e^2}\Big(\tilde{\X}_{Z(t)}^{\prime}\tilde{\X}_{Z(t)}-2\tilde{\X}_{Z(t)}^{\prime}\sum_{j}\B_{ajZ(t)}\alpha_j\mu_{aj}\\
&+2\sum_j\sum_{k\neq j}\B_{ajZ(t)}^{\prime}\B_{akZ(t)}\alpha_j\alpha_k\mu_{aj}\mu_k+\sum_j\B_{ajZ(t)}^{\prime}\B_{ajZ(t)}\big((\mu_{aj}^2+s_j^2)\alpha_j+vs_j^2(1-\alpha_j)\big)\Big)\\
&+\sum_{j=1}^{k_a}\Big(c-\frac{s_j^2}{2\sigma_j^2}+\alpha_j(\ln w_j-\frac{1}{2}\ln\sigma_j^2-\frac{\mu_{aj}^2}{2\sigma_j^2})+(1-\alpha_j)(\ln(1-w_j)-\frac{1}{2}\ln v\sigma_j^2)\Big).
\end{align*}
On the other hand, the joint posterior distribution $q(\btheta_a,\mathbf{r})$ and its logarithm transformation can be expressed as
\begin{align*}
q(\btheta_a,\mathbf{r})&=\prod_{j=1}^{k_a}q(\theta_{aj}|r_j)\prod_{j=1}^{k_a}q(r_j),\\
\ln q(\btheta_a,\mathbf{r})&=\sum_{j=1}^{k_a}\ln q(\theta_{aj}|r_j)+\sum_{j=1}^{k_a}\ln q(r_j).
\end{align*}
The first part of $q(\btheta_a,\mathbf{r})$ and its logarithm transformation can be derived as
\begin{align*}
\prod_{j=1}^{k_a}q(\theta_{aj}|r_j)&=\prod_{j=1}^{k_a}\Big(\frac{1}{\sqrt{2\pi}s_j}\exp(-\frac{(\theta_{aj}-\mu_{aj})^2}{2s_j^2})\Big)^{I(r_j=1)}\Big(\frac{1}{\sqrt{2\pi v}s_j}\exp(-\frac{\theta_{aj}^2}{2vs_j^2})\Big)^{I(r_j=0)},\\
\sum_{j=1}^{k_a}\ln q(\theta_{aj}|r_j)&=\sum_{j=1}^{k_a}(c-\frac{1}{2}\ln s_j^2-\frac{(\theta_{aj}-\mu_{aj})^2}{2s_j^2})r_j+(c-\frac{1}{2}\ln(vs_j^2)-\frac{\theta_{aj}^2}{2vs_j^2})(1-r_j)
\setlength{\belowdisplayskip}{1pt}.
\end{align*}
Take its expectation under the distribution of $q(\btheta_a,\mathbf{r})$
\begin{align*}
E_{q(\btheta_a,\mathbf{r})}\Big(\sum_{j=1}^{k_a}\ln q(\theta_a|r_j)\Big)&=\sum_{j=1}^{k_a}\Big((c-\frac{1}{2}\ln s_j^2-\frac{s_j^2}{2s_j^2})\alpha_j+(c-\frac{1}{2}\ln (vs_j^2)-\frac{vs_j^2}{2vs_j^2})(1-\alpha_j)\Big).
\setlength{\belowdisplayskip}{1pt}
\end{align*}
Also the second part of $q(\btheta_a,\mathbf{r})$ and its logarithm transformation can be derived as
\begin{align*}
\prod_{j=1}^{k_a}q(r_j)&=\prod_{j=1}^{k_a}\alpha_j^{r_j}+(1-\alpha_j)^{1-r_j}, \\
E_{q(\btheta_a,\mathbf{r})}(\sum_{j=1}^{k_a}\ln q(r_j))&=\sum_{j=1}^{k_a}\alpha_j\ln\alpha_j+(1-\alpha_j)\ln(1-\alpha_j).
\setlength{\belowdisplayskip}{1pt}
\end{align*}
Take its expectation under the distribution of $q(\btheta_a,\mathbf{r})$
\begin{align*}
E_{q(\btheta_a,\mathbf{r})}\Big(\ln q(\btheta_a,\mathbf{r})\Big)=\sum_{j=1}^{k_a}\Big(c-\frac{1}{2}+\alpha_j(\ln\alpha_j-\frac{1}{2}\ln s_j^2)+(1-\alpha_j)(\ln(1-\alpha_j)-\frac{1}{2}\ln vs_j^2)\Big)
\end{align*}
To give a summary, the negative KL divergence between the true posterior $p(\tilde{\X},\btheta_a,\mathbf{r})$ and the approximate posterior $q(\btheta_a,\mathbf{r})$ is
\begin{align*}
Z&=E_{q(\btheta_a,\mathbf{r})}\Big(\ln p(\btheta_a,\mathbf{r},\tilde{\X}_{Z(1)},\tilde{\X}_{Z(2)},...\tilde{\X}_{Z(n)})\Big)-E_{q(\btheta_a,\mathbf{r})}\Big(\ln q(\btheta_a,\mathbf{r})\Big)\\
&=pc-\frac{1}{2}\ln\sigma_e^2-\sum_{t=1}^{n}\frac{\lambda_t^n}{2\sigma_e^2}\Big(\tilde{\X}_{Z(t)}^{\prime}\tilde{\X}_{Z(t)}-2\tilde{\X}_{Z(t)}^{\prime}\sum_{j=1}^{k_a}\B_{ajZ(t)}\alpha_j\mu_{aj}+2\sum_j\sum_{k\neq j}\B_{ajZ(t)}^{\prime}\B_{akZ(t)}\alpha_j\alpha_k\mu_{aj}\mu_{ak}\\
&+\sum_{j=1}^{k_a}\B_{ajZ(t)}^{\prime}\B_{ajZ(t)}\big((\mu_{aj}^2+s_j^2)\alpha_j+vs_j^2(1-\alpha_j)\big)\Big)+\sum_{j=1}^{k_a}\Big(\frac{1}{2}-\frac{s_j^2}{2\sigma_j^2}+\frac{1}{2}\ln\frac{s_j^2}{\sigma_j^2}+(\ln w_j-\ln\alpha_j-\frac{\mu_{aj}^2}{2\sigma_j^2})\alpha_j\\
&+(1-\alpha_j)(\ln(1-w_j)-\ln(1-\alpha_j))\Big).
\end{align*}
Taking the partial derivatives of the negative Kullback-Leibler divergence, we obtain the coordinate descent updates for this optimization problem. And let $v \rightarrow 0$, we can obtain 
\begin{flalign*}
&\frac{\partial Z}{\partial s_j^2}=-\sum_{t=1}^{n}\frac{\lambda_t^n}{2\sigma_e^2}\B_{ajZ(t)}^{\prime}\B_{ajZ(t)}(\alpha_j+v(1-\alpha_j))-\frac{\alpha_j}{2\sigma_j^2}+\frac{\alpha_j}{2s_j^2}=0&\\
&s_j^2=\frac{1}{\sum_{t=1}^{n}\frac{\lambda_t^n}{\sigma_e^2}\B_{ajZ(t)}^{\prime}\B_{ajZ(t)}+\frac{1}{\sigma_j^2}}&\\
% &(s_j^2)^{-1}=\sum_{t=1}^{n}\frac{\lambda_t^n}{\sigma_e^2}\B_{ajZ(t)}^{\prime}\B_{ajZ(t)}(1+v(\frac{1}{\alpha_j}-1))+\frac{1}{\sigma_j^2}&\\
&\frac{\partial Z}{\partial \mu_{aj}}=-\sum_{t=1}^{n}\frac{\lambda_t^n}{2\sigma_e^2}(-2\tilde{\X}_{Z(t)}^{\prime}\B_{ajZ(t)}\alpha_j+2\sum_{k \neq j}\B_{ajZ(t)}^{\prime}\B_{akZ(t)}\alpha_j\alpha_k\mu_{ak}+2\B_{ajZ(t)}^{\prime}\B_{ajZ(t)}\alpha_j\mu_{aj})-\frac{\alpha_j\mu_{aj}}{\sigma_j^2}=0&\\
&\mu_{aj}=\frac{s_j^2}{\sigma_e^2}\sum_{t=1}^{n}\lambda_t^n(\tilde{\X}_{Z(t)}^{\prime}\B_{ajZ(t)}-\sum_{k \neq j}\B_{ajZ(t)}^{\prime}\B_{akZ(t)}\alpha_k\mu_{ak})&\\
% &\mu_{aj}=\frac{\sum_{t=1}^{n}\frac{\lambda_t^n}{\sigma_e^2}(\tilde{\X}_{Z(t)}^{\prime}\B_{ajZ(t)}-\sum_{k \neq j}\B_{ajZ(t)}^{\prime}\B_{akZ(t)}\alpha_k\mu_{ak})}{\sum_{t=1}^{n}\frac{\lambda_t^n}{\sigma_e^2}\B_{ajZ(t)}^{\prime}\B_{ajZ(t)}+\frac{1}{\sigma_j^2}}&\\
&\frac{\partial Z}{\partial \alpha_j}=-\sum_{t=1}^{n}\frac{\lambda_t^n}{2\sigma_e^2}\Big(-2\tilde{\X}_{Z(t)}^{\prime}\B_{ajZ(t)}\mu_{aj}+2\sum_{k\neq j}\B_{ajZ(t)}^{\prime}\B_{akZ(t)}\alpha_k\mu_{aj}\mu_{ak}+\B_{ajZ(t)}^{\prime}\B_{ajZ(t)}(\mu_{aj}^2+s_j^2-vs_j^2)\Big)&\\
&+\ln w_j-\frac{\mu_{aj}^2}{2\sigma_j^2}-\ln\alpha_j-\ln(1-w_j)+\ln(1-\alpha_j)=0&\\
&\ln\frac{\alpha_j}{1-\alpha_j}=\ln\frac{w_j}{1-w_j}-\frac{\mu_{aj}^2}{2\sigma_j^2}+\mu_{aj}^2(\sum_{t=1}^{n}\frac{\lambda_t^n}{\sigma_e^2}\B_{ajZ(t)}^{\prime}\B_{ajZ(t)}+\frac{1}{\sigma_j^2})-\sum_{t=1}^{n}\frac{\lambda_t^n}{2\sigma_e^2}\B_{ajZ(t)}^{\prime}\B_{ajZ(t)}(\mu_{aj}^2+s_j^2-vs_j^2)&\\
&=\ln\frac{w_j}{1-w_j}+\frac{\mu_{aj}^2}{2\sigma_j^2}+\sum_{t=1}^{n}\frac{\lambda_t^n}{2\sigma_e^2}\B_{ajZ(t)}^{\prime}\B_{ajZ(t)}(\mu_{aj}^2-s_j^2+vs_j^2)&
\end{flalign*}

\section*{Appendix B: Deviation of Detection Statistic}
\label{Appendix2}
Some notations are defined as 
$\mathbf{A}=\B_{aZ(n)}^{\prime}\bSigma_e^{-1}\B_{aZ(n)}+\mathbf{K}^{-1}$, $\mathbf{D}=\X_{Z(n)}^{\prime}\bSigma_e^{-1}\B_{aZ(n)}+\bmu_r'\mathbf{K}^{-1}$, $\mathbf{C}=\B_{bZ(n)}^{\prime}\bSigma_e^{-1}\B_{aZ(n)}$, 
$\mathbf{H}=\B_{bZ(n)}^{\prime}\bSigma_e^{-1}\B_{bZ(n)}+\tilde{\bSigma}_b^{-1}$,$\mathbf{G}^{[0]}=\X_{Z(n)}^{\prime}\bSigma_e^{-1}\B_{bZ(n)}+\tilde{\btheta}_n^{[0]\prime}\tilde{\bSigma}_b^{-1}$,$\mathbf{G}^{[1]}=\X_{Z(n)}^{\prime}\bSigma_e^{-1}\B_{bZ(n)}+\tilde{\btheta}_n^{[1]\prime}\tilde{\bSigma}_b^{-1}$.

The marginal likelihood under $H_0$, which is the numerator of the Bayesian factor, can be derived as
\begin{flalign*}
&P(\X_{Z(n)}|H_0)=\int \int p(\btheta_a|H_0)p(\btheta_n|\btheta_a,H_0)p(\X_{Z(n)}|\btheta_a,\btheta_n)d\btheta_nd\btheta_a&\\
&=\sqrt{1/\big((2\pi)^{k_b+m}|\tilde{\bSigma}_b||\bSigma_e|\big)}\int \exp\bigg(-\frac{1}{2}\Big((\btheta_n-\tilde{\btheta}_n^{[0]})^{\prime}\tilde{\bSigma}_b^{-1}(\btheta_n-\tilde{\btheta}_n)+(\X_{Z(n)}-\B_{bZ(n)}\btheta_n)^{\prime}\bSigma_e^{-1}&\\
&(\X_{Z(n)}-\B_{bZ(n)}\btheta_n)\Big)\bigg)d\btheta_n&\\
&=\sqrt{1/\big((2\pi)^{k_b+m}|\tilde{\bSigma}_b||\bSigma_e|\big)}\int \exp\bigg(-\frac{1}{2}\Big(\btheta_n^{\prime}(\tilde{\bSigma}_b^{-1}+\B_{bZ(n)}^{\prime}\bSigma_e^{-1}\B_{bZ(n)})\btheta_n-2(\X_{Z(n)}'\bSigma_e^{-1}\B_{bZ(n)}+\tilde{\btheta}_n^{[0]\prime}\tilde{\bSigma}_b^{-1})&\\
&\btheta_n+\tilde{\btheta}_n^{[0]\prime}\tilde{\bSigma}_b^{-1}\tilde{\btheta}_n^{[0]}+\X_{Z(n)}^{\prime}\bSigma_e^{-1}\X_{Z(n)}\Big)\bigg)d\btheta_n&\\
&=\sqrt{1/\big((2\pi)^{k_b+m}|\tilde{\bSigma}_b||\bSigma_e|\big)}\exp\Big(-\frac{1}{2}(\tilde{\btheta}_n^{[0]\prime}\tilde{\bSigma}_b^{-1}\tilde{\btheta}_n^{[0]}+\X_{Z(n)}^{\prime}\bSigma_0^{-1}\X_{Z(n)})\Big) \int \exp\Big(\frac{1}{2}((\btheta_n-\mathbf{H}^{-1}\mathbf{G}^{[0]\prime})^{\prime}\mathbf{H}&\\
&(\btheta_n-\mathbf{H}^{-1}\mathbf{G}^{[0]\prime})-\mathbf{G}^{[0]}\mathbf{H}^{-1}\mathbf{G}^{[0]\prime})\Big)d\btheta_n&\\
&=\sqrt{1/\big((2\pi)^{m}|\tilde{\bSigma}_b||\bSigma_e||\mathbf{H}|\big)}\exp\Big(-\frac{1}{2}(\tilde{\btheta}_n^{[0]\prime}\tilde{\bSigma}_b^{-1}\tilde{\btheta}_n^{[0]}+\X_{Z(n)}^{\prime}\bSigma_e^{-1}\X_{Z(n)}-\mathbf{G}^{[0]}\mathbf{H}^{-1}\mathbf{G}^{[0]\prime})\Big).&
\end{flalign*}
The marginal likelihood under $H_1$, which is the denominator of the Bayesian factor, can be derived as
\begin{flalign*}
&P(X_{Z(n)}|H_1)=\sum_{r}\int \int p(\mathbf{r}|H_1)p(\btheta_a|\mathbf{r},H_1)p(\btheta_n|\btheta_a,H_1)p(\X_{Z(n)}|\btheta_a,\btheta_n)d\btheta_nd\btheta_a&\\
&=\sum_{\mathbf{r}}p(\mathbf{r}|H_1)\sqrt{1/\big((2\pi)^{k_a+k_b+m}|\mathbf{K}||\tilde{\bSigma}_b||\bSigma_e|\big)}\int \int p(\btheta_a|\mathbf{r},H_1)p(\btheta_n|\btheta_a,H_1)p(\X_{Z(n)}|\btheta_a,\btheta_n)d\btheta_nd\btheta_a&\\
&=\sum_{\mathbf{r}}p(\mathbf{r}|H_1)\sqrt{1/\big((2\pi)^{k_a+k_b+m}|\mathbf{K}||\tilde{\bSigma}_b||\bSigma_e|\big)}\int \int \exp\bigg(-\frac{1}{2}\Big((\btheta_a-\bmu_r)^{\prime}\mathbf{K}^{-1}(\btheta_a-\bmu_r)+(\btheta_n-\tilde{\btheta}_n)^{[1]\prime}&\\
&\tilde{\bSigma}_b^{-1}(\btheta_n-\tilde{\btheta}_n^{[1]})+(\X_{Z(n)}-\B_{bZ(n)}\btheta_n-\B_{aZ(n)}\btheta_a)^{\prime}\bSigma_e^{-1}(\X_{Z(n)}-\B_{bZ(n)}\btheta_n-\B_{aZ(n)}\btheta_a)\Big)\bigg)d\btheta_ad\btheta_n&\\
&=\sum_{\mathbf{r}}p(\mathbf{r}|H_1)\sqrt{1/\big((2\pi)^{k_a+k_b+m}|\mathbf{K}||\tilde{\bSigma}_b||\bSigma_e|\big)}\int \int \exp\bigg(-\frac{1}{2}\Big(\btheta_a^{\prime}(\B_{aZ(n)}^{\prime}\bSigma_e^{-1}\B_{aZ(n)}+\mathbf{K}^{-1})\btheta_a&\\
&-2((\X_{Z(n)}-\B_{bZ(n)}\btheta_n)^{\prime}\bSigma_e^{-1}\B_{aZ(n)}+\bmu_r^{\prime}\mathbf{K}^{-1})\btheta_a+\bmu_r^{\prime}\mathbf{K}^{-1}\bmu_r+\btheta_n^{\prime}(\B_{bZ(n)}^{\prime}\bSigma_e^{-1}\B_{bZ(n)}+\tilde{\bSigma}_b^{-1})\btheta_n&\\
&-2(\X_{Z(n)}^{\prime}\bSigma_e^{-1}\B_{bZ(n)}+\tilde{\btheta}_n^{[1]\prime}\tilde{\bSigma}_b^{-1})\btheta_n+\X_{Z(n)}^{\prime}\bSigma_e^{-1}\X_{Z(n)}+\tilde{\btheta}_n^{[1]\prime}\tilde{\bSigma}_b^{-1}\tilde{\btheta}_n^{[1]}\Big)\bigg)d\btheta_ad\btheta_n&\\
&=\sum_{\mathbf{r}}p(\mathbf{r}|H_1)\sqrt{1/\big((2\pi)^{k_a+k_b+m}|\mathbf{K}||\tilde{\bSigma}_b||\bSigma_e|\big)}\int \int \exp\bigg(-\frac{1}{2}\Big(\btheta_a^{\prime}\mathbf{A}\btheta_a-2(\mathbf{R}-\btheta_n^{\prime}\mathbf{C})\btheta_a+\btheta_n^{\prime}\mathbf{H}\btheta_n&\\
&-2\mathbf{G}^{[1]}\btheta_n+\bmu_r^{\prime}\mathbf{K}^{-1}\bmu_r+\X_{Z(n)}^{\prime}\bSigma_e^{-1}\X_{Z(n)}+\tilde{\btheta}_n^{[1]\prime}\tilde{\bSigma}_b^{-1}\tilde{\btheta}_n^{[1]}\Big)\bigg)d\btheta_ad\btheta_n&\\
&=\sum_{\mathbf{r}}p(\mathbf{r}|H_1)\sqrt{1/\big((2\pi)^{k_a+k_b+m}|\mathbf{K}||\tilde{\bSigma}_b||\bSigma_e|\big)}\exp\Big(-\frac{1}{2}(\bmu_r^{\prime}\mathbf{K}^{-1}\bmu_r+\X_{Z(n)}^{\prime}\bSigma_e^{-1}\X_{Z(n)}+\tilde{\btheta}_n^{[1]\prime}\tilde{\bSigma}_b^{-1}\tilde{\btheta}_n^{[1]})\Big)\\
&\sqrt{(2\pi)^{k_a}/|\mathbf{A}|}\int \exp\Big(-\frac{1}{2}(\btheta_n^{\prime}(\mathbf{H}-\mathbf{C}\mathbf{A}^{-1}\mathbf{C}^{\prime})\btheta_n-2(\mathbf{G}^{[1]}-\mathbf{R}\mathbf{A}^{-1}\mathbf{C}^{\prime})\btheta_n-\mathbf{R}\mathbf{A}^{-1}\mathbf{R}^{\prime})\Big)d\btheta_n\\
&=\sum_{\mathbf{r}}p(\mathbf{r}|H_1)\sqrt{1/\big((2\pi)^{k_b+m}|\mathbf{K}||\tilde{\bSigma}_b||\bSigma_e||\mathbf{A}|\big)}\exp\Big(-\frac{1}{2}(\bmu_r^{\prime}\mathbf{K}^{-1}\bmu_r+\X_{Z(n)}^{\prime}\bSigma_e^{-1}\X_{Z(n)}+\tilde{\btheta}_n^{[1]\prime}\tilde{\bSigma}_b^{-1}\tilde{\btheta}_n^{[1]})\Big)&\\
&\sqrt{(2\pi)^{k_b}/|\mathbf{H}-\mathbf{C}\mathbf{A}^{-1}\mathbf{C}^{\prime}|}\exp\Big(-\frac{1}{2}(-\mathbf{R}\mathbf{A}^{-1}\mathbf{R}^{\prime}-(\mathbf{G}^{[1]}-\mathbf{R}\mathbf{A}^{-1}\mathbf{C}^{\prime})(\mathbf{H}-\mathbf{C}\mathbf{A}^{-1}\mathbf{C}^{\prime})^{-1}(\mathbf{G}^{[1]}-\mathbf{R}\mathbf{A}^{-1}\mathbf{C}^{\prime})^{\prime})\Big)&\\
&=\sum_{\mathbf{r}}p(\mathbf{r}|H_1)\sqrt{1/\big((2\pi)^{m}|\mathbf{K}||\tilde{\bSigma}_b||\bSigma_e||\mathbf{A}||\mathbf{H}-\mathbf{C}\mathbf{A}^{-1}\mathbf{C}^{\prime}|\big)}\exp\left\{-\frac{1}{2}(\bmu_r^{\prime}\mathbf{K}^{-1}\bmu_r+\X_{Z(n)}^{\prime}\bSigma_e^{-1}\X_{Z(n)}\right.&\\
&\left.+\tilde{\btheta}_n^{[1]\prime}\tilde{\bSigma}_b^{-1}\tilde{\btheta}_n^{[1]}-\mathbf{R}\mathbf{A}^{-1}\mathbf{R}^{\prime}-(\mathbf{G}^{[1]}-\mathbf{R}\mathbf{A}^{-1}\mathbf{C}^{\prime})(\mathbf{H}-\mathbf{C}\mathbf{A}^{-1}\mathbf{C}^{\prime})^{-1}(\mathbf{G}^{[1]}-\mathbf{R}\mathbf{A}^{-1}\mathbf{C}^{\prime})^{\prime})\right\}.&
\end{flalign*}
So the posterior Bayesian factor is derived as
\begin{align*}
&PBF_n=\frac{\sum_{\mathbf{r}}\int \int p(\mathbf{r}|H_1) p(\btheta_a|H_1)p(\btheta_n|\btheta_a,H_1)p(\X_{Z(n)}|\btheta_a,\btheta_n)d\btheta_nd\btheta_a}{\int \int p(\btheta_a|H_0)p(\btheta_n|\btheta_a,H_0)p(\X_{Z(n)}|\btheta_a,\btheta_n)d\btheta_nd\btheta_a}\\
&=\sum_{\mathbf{r}}p(\mathbf{r}|H_1)\sqrt{|\mathbf{H}|/\Big(|\mathbf{K}||\mathbf{A}||\mathbf{H}-\mathbf{C}\mathbf{A}^{-1}\mathbf{C}^{\prime}|\Big)}\exp\bigg(-\frac{1}{2}\Big(\bmu_r^{\prime}\mathbf{K}^{-1}\bmu_r+\mathbf{G}^{[0]}\mathbf{H}^{-1}\mathbf{G}^{[0]\prime}-\mathbf{R}\mathbf{A}^{-1}\mathbf{R}^{\prime}\\
&-(\mathbf{G}^{[1]}-\mathbf{R}\mathbf{A}^{-1}\mathbf{C}^{\prime})(\mathbf{H}-\mathbf{C}\mathbf{A}^{-1}\mathbf{C}^{\prime})^{-1}(\mathbf{G}^{[1]}-\mathbf{R}\mathbf{A}^{-1}\mathbf{C}^{\prime})^{\prime}+\tilde{\btheta}_n^{[1]\prime}\tilde{\bSigma}_b^{-1}\tilde{\btheta}_n^{[1]}-\tilde{\btheta}_n^{[0]\prime}\tilde{\bSigma}_b^{-1}\tilde{\btheta}_n^{[0]}\Big)\bigg)
\setlength{\belowdisplayskip}{1pt}.
\end{align*}
Note that the form of $PBF_n$ is too complex for easy computation. We consider further simplifying it by eliminating constants and small values as below.

% Under the assumptions that $\B_{aZ(n)}^{\prime}\B_{aZ(n)}$ is nearly an unitary matrix, spectral radius $\rho(\mathbf{K}\B_{aZ(n)}^{\prime}\B_{aZ(n)})<1$
% and $\mathbf{K}^{-1}$ is nearly full-rank diagonal matrix, then
For a square matrix $\mathbf{M}$ with spectral radius $\rho(\mathbf{M})<1$, according to the Maclaurin series of  matrix form, $(\mathbf{I}+\mathbf{M})^{-1}=\sum_{k=0}^{\infty}(-1)^k\mathbf{M}^k$.
Since $\mathbf{K}\frac{\B_{aZ(n)}^{\prime}\B_{aZ(n)}}{\sigma_e^2}$ is a square matrix and the entries of $\mathbf{K}$ are quite small, the spectral radius $\rho(\mathbf{K}\frac{\B_{aZ(n)}^{\prime}\B_{aZ(n)}}{\sigma_e^2})\leq \Vert\mathbf{K}\frac{\B_{aZ(n)}^{\prime}\B_{aZ(n)}}{\sigma_e^2}\Vert<1$ can be satisfied. Then we can generalize it as
\begin{align*}
&\mathbf{A}^{-1}=(\frac{\B_{aZ(n)}^{\prime}\B_{aZ(n)}}{\sigma_e^2}+\mathbf{K}^{-1})^{-1}=\sum_{k=0}^{\infty}(-1)^k(\mathbf{K}\frac{\B_{aZ(n)}^{\prime}\B_{aZ(n)}}{\sigma_e^2})^k\mathbf{K}\approx \mathbf{K}-\mathbf{K}\frac{\B_{aZ(n)}^{\prime}\B_{aZ(n)}}{\sigma_e^2}\mathbf{K}+o(\mathbf{K}^2).
\end{align*}
Following the same way, consider $\mathbf{H}^{-1}\mathbf{C}\mathbf{A}^{-1}\mathbf{C}^{\prime}$ is a square matrix and the entries of $\mathbf{H}^{-1}$ are quite small. The spectral radius $\rho(\mathbf{H}^{-1}\mathbf{C}\mathbf{A}^{-1}\mathbf{C}^{\prime})\leq \Vert\mathbf{H}^{-1}\mathbf{C}\mathbf{A}^{-1}\mathbf{C}^{\prime}\Vert<1$ can be satisfied as well. Then,
\begin{align*}
&(\mathbf{H}-\mathbf{C}\mathbf{A}^{-1}\mathbf{C}^{\prime})^{-1}=\mathbf{H}^{-1}+\mathbf{H}^{-1}\mathbf{C}\mathbf{A}^{-1}\mathbf{C}^{\prime}\mathbf{H}^{-1}+o((\mathbf{H}^{-1})^2).
\end{align*}
Then the following items can be simplified as
\begin{align*}
&(\mathbf{G}^{[1]}-\mathbf{R}\mathbf{A}^{-1}\mathbf{C}^{\prime})(\mathbf{H}-\mathbf{C}\mathbf{A}^{-1}\mathbf{C}^{\prime})^{-1}(\mathbf{G}^{[1]}-\mathbf{R}\mathbf{A}^{-1}\mathbf{C}^{\prime})^{\prime}=(\mathbf{G}^{[1]}-\mathbf{R}\mathbf{A}^{-1}\mathbf{C}^{\prime})\mathbf{H}^{-1}(\mathbf{G}^{[1]}-\mathbf{R}\mathbf{A}^{-1}\mathbf{C}^{\prime})^{\prime}\\
&+(\mathbf{G}^{[1]}-\mathbf{R}\mathbf{A}^{-1}\mathbf{C}^{\prime})\mathbf{H}^{-1}\mathbf{C}\mathbf{A}^{-1}\mathbf{C}^{\prime}\mathbf{H}^{-1}(\mathbf{G}^{[1]}-\mathbf{R}\mathbf{A}^{-1}\mathbf{C}^{\prime})^{\prime}\\
&=\mathbf{G}^{[1]}\mathbf{H}^{-1}\mathbf{G}^{[1]\prime}+(\mathbf{R}\mathbf{A}^{-1}\mathbf{C}^{\prime})\mathbf{H}^{-1}(\mathbf{R}\mathbf{A}^{-1}\mathbf{C}^{\prime})^{\prime}-2\mathbf{R}\mathbf{A}^{-1}\mathbf{C}^{\prime}\mathbf{H}^{-1}\mathbf{G}^{[1]\prime}.\\
&\mathbf{R}\mathbf{A}^{-1}\mathbf{C}^{\prime}=(\frac{\X_{Z(n)}^{\prime}\B_{aZ(n)}}{\sigma_e^2}+\bmu_r^{\prime}\mathbf{K}^{-1})\mathbf{K}\frac{\B_{aZ(n)}^{\prime}\B_{bZ(n)}}{\sigma_e^2}=\frac{\X_{Z(n)}^{\prime}\B_{aZ(n)}}{\sigma_e^2}\mathbf{K}\frac{\B_{aZ(n)}^{\prime}\B_{bZ(n)}}{\sigma_e^2}+\bmu_r^{\prime}\frac{\B_{aZ(n)}^{\prime}\B_{bZ(n)}}{\sigma_e^2}.\\
&\mathbf{R}\mathbf{A}^{-1}\mathbf{R}^{\prime}=(\frac{\X_{Z(n)}^{\prime}\B_{aZ(n)}}{\sigma_e^2}+\bmu_r^{\prime}\mathbf{K}^{-1})(\mathbf{K}-\mathbf{K}\frac{\B_{aZ(n)}^{\prime}\B_{aZ(n)}}{\sigma_e^2}\mathbf{K})(\frac{\X_{Z(n)}^{\prime}\B_{aZ(n)}}{\sigma_e^2}+\bmu_r^{\prime}\mathbf{K}^{-1})^{\prime}\\
&=\frac{\X_{Z(n)}^{\prime}\B_{aZ(n)}}{\sigma_e^2}\mathbf{K}\frac{\B_{aZ(n)}^{\prime}\X_{Z(n)}}{\sigma_e^2}+\bmu_r^{\prime}\mathbf{K}^{-1}\bmu_r+2\frac{\X_{Z(n)}^{\prime}\B_{aZ(n)}}{\sigma_e^2}\bmu_r-\bmu_r^{\prime}\frac{\B_{aZ(n)}^{\prime}\B_{aZ(n)}}{\sigma_e^2}\bmu_r\\
&-2\bmu_r^{\prime}\frac{\B_{aZ(n)}^{\prime}\B_{aZ(n)}}{\sigma_e^2}\mathbf{K}\frac{\B_{aZ(n)}^{\prime}\X_{Z(n)}}{\sigma_e^2}\\
&=\bmu_r^{\prime}\mathbf{K}^{-1}\bmu_r+2\frac{\X_{Z(n)}^{\prime}\B_{aZ(n)}}{\sigma_e^2}\bmu_r-\bmu_r^{\prime}\frac{\B_{aZ(n)}^{\prime}\B_{aZ(n)}}{\sigma_e^2}\bmu_r+o(\mathbf{K})
\setlength{\belowdisplayskip}{1pt}.
\end{align*}
% Under the assumptions that $\B_{bZ(n)}^{\prime}\B_{bZ(n)}$ is nearly an unitary matrix, spectral radius $\rho(\tilde{\bSigma}_b\B_{bZ(n)}^{\prime}\B_{bZ(n)})<1$
% and $\tilde{\bSigma}_b^{-1}$ is nearly full-rank diagonal matrix, then
Similarly, consider  $\tilde{\bSigma}_b\frac{{\B_{bZ(n)}}^{\prime}\B_{bZ(n)}}{\sigma_e^2}$ is a square matrix and the entries of $\tilde{\bSigma}_b$ are quite small. Its spectral radius $\rho(\tilde{\bSigma}_b\frac{{\B_{bZ(n)}}^{\prime}\B_{bZ(n)}}{\sigma_e^2})\leq \Vert\tilde{\bSigma}_b\frac{{\B_{bZ(n)}}^{\prime}\B_{bZ(n)}}{\sigma_e^2}\Vert<1$ can be satisfied. Then,
\begin{flalign*}
&\mathbf{H}^{-1}=(\frac{{\B_{bZ(n)}}^{\prime}\B_{bZ(n)}}{\sigma_e^2}+\tilde{\bSigma}_b^{-1})^{-1}=\tilde{\bSigma}_b-\tilde{\bSigma}_b\frac{{\B_{bZ(n)}}^{\prime}\B_{bZ(n)}}{\sigma_e^2}\tilde{\bSigma}_b+\tilde{\bSigma}_b\frac{\B_{bZ(n)}^{\prime}\B_{bZ(n)}}{\sigma_e^2}\tilde{\bSigma}_b\frac{{\B_{bZ(n)}}^{\prime}\B_{bZ(n)}}{\sigma_e^2}\tilde{\bSigma}_b&\\
&+o(\tilde{\bSigma}_b^2).&\\
&\mathbf{H}^{-1}\mathbf{G}^{[1]\prime}=(\tilde{\bSigma}_b-\tilde{\bSigma}_b\frac{{\B_{bZ(n)}}^{\prime}\B_{bZ(n)}}{\sigma_e^2}\tilde{\bSigma}_b)(\tilde{\bSigma}_b^{-1}\tilde{\btheta}_{n}^{[1]}+\frac{{\B_{bZ(n)}}^{\prime}\X_{Z(n)}}{\sigma_e^2})&\\
&=\tilde{\btheta}_{n}^{[1]}-\tilde{\bSigma}_b\frac{\B_{bZ(n)}^{\prime}\B_{bZ(n)}}{\sigma_e^2}\tilde{\btheta}_{n}^{[1]}-\tilde{\bSigma}_b\frac{{\B_{bZ(n)}}^{\prime}\B_{bZ(n)}}{\sigma_e^2}\tilde{\bSigma}_b\frac{\B_{bZ(n)}^{\prime}\X_{Z(n)}}{\sigma_e^2}+\tilde{\bSigma}_b\frac{\B_{bZ(n)}^{\prime}\X_{Z(n)}}{\sigma_e^2}.&\\
&\mathbf{R}\mathbf{A}^{-1}\mathbf{C}^{\prime}\mathbf{H}^{-1}\mathbf{
G}^{\prime}=\bmu_r\frac{\B_{aZ(n)}^{\prime}\B_{aZ(n)}}{\sigma_e^2}\tilde{\btheta}_{n}^{[1]}-\bmu_r^{\prime}\frac{\B_{aZ(n)}^{\prime}\B_{bZ(n)}}{\sigma_e^2}\tilde{\bSigma}_b\frac{\B_{bZ(n)}^{\prime}\B_{bZ(n)}}{\sigma_e^2}\tilde{\btheta}_{n}^{[1]}&\\
&-\bmu_r^{\prime}\frac{\B_{aZ(n)}^{\prime}\B_{bZ(n)}}{\sigma_e^2}\tilde{\bSigma}_b\frac{\B_{bZ(n)}^{\prime}\B_{bZ(n)}}{\sigma_e^2}\tilde{\bSigma}_b\frac{\B_{bZ(n)}^{\prime}\X_{Z(n)}}{\sigma_e^2}+\bmu_r^{\prime}\frac{\B_{aZ(n)}^{\prime}\B_{bZ(n)}}{\sigma_e^2}\tilde{\bSigma}_b\frac{\B_{bZ(n)}^{\prime}\X_{Z(n)}}{\sigma_e^2}.&\\
&\mathbf{R}\mathbf{A}^{-1}\mathbf{C}^{\prime}\mathbf{H}^{-1}(\mathbf{R}\mathbf{A}^{-1}\mathbf{C}^{\prime})^{\prime}=\bmu_r^{\prime}\frac{\B_{aZ(n)}^{\prime}\B_{bZ(n)}}{\sigma_e^2}\tilde{\bSigma}_b\frac{\B_{bZ(n)}^{\prime}\B_{aZ(n)}}{\sigma_e^2}\bmu_r-\bmu_r^{\prime}\frac{\B_{aZ(n)}^{\prime}\B_{bZ(n)}}{\sigma_e^2}\tilde{\bSigma}_b\frac{\B_{bZ(n)}^{\prime}\B_{bZ(n)}}{\sigma_e^2}&\\
&\tilde{\bSigma}_b\frac{\B_{bZ(n)}^{\prime}\B_{aZ(n)}}{\sigma_e^2}\bmu_r.&
\setlength{\belowdisplayskip}{1pt}
\end{flalign*}
So the terms inside the exponential function in $PBF_n$ is simplified as
\begin{flalign*}
&\bmu_r^{\prime}\mathbf{K}^{-1}\bmu_r+\mathbf{G}^{[0]}\mathbf{H}^{-1}\mathbf{G}^{[0]\prime}-\mathbf{R}\mathbf{A}^{-1}\mathbf{R}^{\prime}-(\mathbf{G}^{[1]}-\mathbf{R}\mathbf{A}^{-1}\mathbf{C}^{\prime})(\mathbf{H}-\mathbf{C}\mathbf{A}^{-1}\mathbf{C}^{\prime})^{-1}(\mathbf{G}^{[1]}-\mathbf{R}\mathbf{A}^{-1}\mathbf{C}^{\prime})^{\prime}&\\
&+\tilde{\btheta}_{n}^{[1]\prime}\tilde{\bSigma}_b^{-1}\tilde{\btheta}_{n}^{[1]}-\tilde{\btheta}_{n}^{[0]\prime}\tilde{\bSigma}_b^{-1}\tilde{\btheta}_{n}^{[0]}&\\
&=\bmu_r^{\prime}\mathbf{K}^{-1}\bmu_r+\mathbf{G}^{[0]}\mathbf{H}^{-1}\mathbf{G}^{[0]\prime}-\mathbf{R}\mathbf{A}^{-1}\mathbf{R}^{\prime}-\mathbf{G}^{[1]}\mathbf{H}^{-1}\mathbf{G}^{[1]\prime}-(\mathbf{R}\mathbf{A}^{-1}\mathbf{C}^{\prime})\mathbf{H}^{-1}(\mathbf{R}\mathbf{A}^{-1}\mathbf{C}^{\prime})^{\prime}&\\
&+2\mathbf{R}\mathbf{A}^{-1}\mathbf{C}^{\prime}\mathbf{H}^{-1}\mathbf{G}^{[1]\prime}+\tilde{\btheta}_{n}^{[1]\prime}\tilde{\bSigma}_b^{-1}\tilde{\btheta}_{n}^{[1]}-\tilde{\btheta}_{n}^{[0]\prime}\tilde{\bSigma}_b^{-1}\tilde{\btheta}_{n}^{[0]}&\\
&=2\frac{\X_{Z(n)}^{\prime}\B_{bZ(n)}}{\sigma_e^2}\tilde{\bSigma}_b\frac{\B_{bZ(n)}^{\prime}\B_{aZ(n)}}{\sigma_e^2}\tilde{\bmu}_a-\tilde{\bmu}_a^{\prime}\frac{\B_{aZ(n)}^{\prime}\B_{bZ(n)}}{\sigma_e^2}\tilde{\bSigma}_b\frac{\B_{bZ(n)}^{\prime}\B_{bZ(n)}}{\sigma_e^2}\tilde{\bSigma}_b\frac{\B_{bZ(n)}^{\prime}\B_{aZ(n)}}{\sigma_e^2}\tilde{\bmu}_a&\\
&-2\frac{\X_{Z(n)}^{\prime}\B_{aZ(n)}}{\sigma_e^2}\bmu_r+\bmu_r^{\prime}\frac{\B_{aZ(n)}^{\prime}\B_{aZ(n)}}{\sigma_e^2}\bmu_r-\bmu_r^{\prime}\frac{\B_{aZ(n)}^{\prime}\B_{bZ(n)}}{\sigma_e^2}\tilde{\bSigma}_b\frac{\B_{bZ(n)}^{\prime}\B_{aZ(n)}}{\sigma_e^2}\bmu_r&\\
&+\bmu_r^{\prime}\frac{\B_{aZ(n)}^{\prime}\B_{bZ(n)}}{\sigma_e^2}\tilde{\bSigma}_b\frac{\B_{bZ(n)}^{\prime}\B_{bZ(n)}}{\sigma_e^2}\tilde{\bSigma}_b\frac{\B_{bZ(n)}^{\prime}\B_{aZ(n)}}{\sigma_e^2}\bmu_r+2\bmu_r^{\prime}\frac{\B_{aZ(n)}^{\prime}\B_{bZ(n)}}{\sigma_e^2}\tilde{\btheta}_{n}^{[1]}&\\
&-2\bmu_r^{\prime}\frac{\B_{aZ(n)}^{\prime}\B_{bZ(n)}}{\sigma_e^2}\tilde{\bSigma}_b\frac{\B_{bZ(n)}^{\prime}\B_{bZ(n)}}{\sigma_e^2}\tilde{\btheta}_{n}^{[1]}+2\bmu_r^{\prime}\frac{\B_{aZ(n)}^{\prime}\B_{bZ(n)}}{\sigma_e^2}\tilde{\bSigma}_b\frac{\B_{bZ(n)}^{\prime}\X_{Z(n)}}{\sigma_e^2}&\\
&-2\bmu_r^{\prime}\frac{\B_{aZ(n)}^{\prime}\B_{bZ(n)}}{\sigma_e^2}\tilde{\bSigma}_b\frac{\B_{bZ(n)}^{\prime}\B_{bZ(n)}}{\sigma_e^2}\tilde{\bSigma}_b\frac{\B_{aZ(n)}^{\prime}\X_{Z(n)}}{\sigma_e^2}&\\
&=-2\bmu_r\frac{\B_{aZ(n)}^{\prime}}{\sigma_e^2}(\X_{Z(n)}-\B_{bZ(n)}\tilde{\btheta}_{n}^{[1]})+\bmu_r^{\prime}\frac{\B_{aZ(n)}^{\prime}\B_{aZ(n)}}{\sigma_e^2}\bmu_r+2\tilde{\bmu}_a^{\prime}\frac{\B_{aZ(n)}^{\prime}}{\sigma_e^2}\hat{\mathbf{H}}\X_{Z(n)}-2\bmu_r^{\prime}\frac{\B_{aZ(n)}^{\prime}}{\sigma_e^2}\hat{\mathbf{H}}\B_{bZ(n)}\tilde{\btheta}_{n}^{[1]}&\\
&-\tilde{\bmu}_a^{\prime}\frac{\B_{aZ(n)}^{\prime}}{\sigma_e^2}\B_{aZ(n)}\tilde{\bmu}_a,&
\end{flalign*}
where $\hat{\mathbf{H}}=\B_{bZ(n)}(\B_{bZ(n)}^{\prime}\B_{bZ(n)})^{-1}\B_{bZ(n)}^{\prime}$. Since $\mathbf{C}\mathbf{A}^{-1}\mathbf{C}^{\prime}$ is quite small and $\mathbf{A}^{-1} \approx \mathbf{K}$, $\sqrt{|\mathbf{H}|/|\mathbf{K}||\mathbf{A}||\mathbf{H}-\mathbf{C}\mathbf{A}^{-1}\mathbf{C}^{\prime}|}\approx 1$, we drop this constant term.
Furthermore, we consider Taylor expansion on each exponential term of $\Lambda_n$, i.e., $\exp(x)=1+x+x^2/2!+x^3/3!+\cdots$ for futher simplification. Consider under $H_{0}$, the term inside the exponential function is usually very close to zero. The first-order expansion would be a sufficient approximation. So the detection statistic can be defined as
\begin{flalign*}
&\Lambda_n \equiv 2\tilde{\bmu}_a^{\prime}\B_{aZ(n)}^{\prime}(\mathbf{I}-\hat{\mathbf{H}})(\X_{Z(n)}-\B_{bZ(n)}\tilde{\btheta}_n^{[1]})-\bmu_a^{\prime}(\B_{aZ(n)}^{\prime}\B_{aZ(n)}\circ\bar{\mathbf{A}})\bmu_a+\tilde{\bmu}_a^{\prime}\B_{aZ(n)}^{\prime}\hat{\mathbf{H}}\B_{aZ(n)}\tilde{\bmu}_a,
\end{flalign*}
where $\bar{\bA}$ has diagonal items $\bar{A}_{ii}=\alpha_{i}, i = 1,\ldots, k_a$, and other items $\bar{A}_{ij}=\alpha_{i}\alpha_{j}, \forall i, j = 1,\ldots, k_a, i\neq j$.
\section*{Appendix C: Verification of Subspace Orthogonal Property}
% Lemma(Hoeffding's inequality). Let $X_1,X_2,...,X_k$ be independent random variables such that, $\forall i \in 1,2,...,k$ we have $X_i \in [a_i,b_i]$ with probability $1$. Denote by $S_k:=\sum_{i=1}^k X_i$ and fix $t>0$. Then:
% \begin{equation*}
% Pr\{|S_k-E[S_k]|\geq t\}\leq 2exp\Big(-\frac{2t^2}{\sum_{i=1}^k(b_i-a_i)^2}\Big)
% \end{equation*}
For a vector $\bb \in \mathbb{R}^p$ with $\Vert\bb\Vert_2^2=1$, denote $\bb^2=(b_1^2,b_2^2,...,b_p^2)^{\prime}$ and assume that $\Vert \bb^2 \Vert_{\infty} \leq \frac{c}{p}\Vert \bb \Vert_2^2$, where $1 \leq c \leq p$. Consider $\mathbf{P}$ is subspace projection matrix from $\mathbb{R}^p \mapsto \mathbb{R}^m$, where $m$ out of $p$ dimensions have $P_{ii}=1$, and all other entries of $\mathbf{P}$ have values of $0$.
Without loss of generality, we can assume that $E[P_{ii}] \geq E[P_{jj}]$ for any $i \leq j$. Thus, $E[\mathbf{P}]=\tilde{\mathbf{P}}=diag\{a_1,a_2,...,a_p\}$,
% \begin{equation*}
% E[P]=\tilde{P}=\left[
% \begin{matrix}
% a_1   &     &  &      \\
% & a_2 &  &       \\
% &     & \ddots & \\
% &     &  & a_p    \\
% \end{matrix}
% \right]
% \end{equation*}
where $0 \leq a_p \leq a_{p-1} \leq ... \leq a_1 \leq 1$ and $\sum_{i=1}^p a_i=m$. Then we have
\begin{equation*}
a_p\lVert \bb \rVert_2^2 \leq E[\lVert \mathbf{P}\bb \rVert_2^2]=\mathbf{\tilde{P}}\lVert \bb \rVert_2^2 \leq a_1\lVert \bb \rVert_2^2.
\end{equation*}
Following \citet{hoeffding1994probability,dasgupta1999elementary}, we have the one side
\begin{align*}
Pr\{\frac{1}{a_1}\lVert \mathbf{P}\bb \rVert_2^2-\lVert \bb \rVert_2^2 \geq \epsilon\}
\leq Pr\{\frac{1}{a_1}\lVert \mathbf{P}\bb \rVert_2^2-\frac{1}{a_1}E[\lVert \mathbf{P}\bb \rVert_2^2] \geq \epsilon\}
\leq exp\Big(-\frac{2a_1^2\epsilon^2}{m\lVert \bb^2 \rVert_\infty^2}\Big)\\
\leq exp\Big(-\frac{2a_1^2p^2\epsilon^2}{mc^2}\Big)\leq exp\Big(-\frac{2(\frac{m}{p})^2p^2\epsilon^2}{mc^2}\Big)=exp\Big(-\frac{2m\epsilon^2}{c^2}\Big),
\end{align*}
and the other side 
\begin{align*}
Pr\{\lVert \bb \rVert_2^2-\frac{1}{a_p}\lVert \mathbf{P}\bb \rVert_2^2 \geq \epsilon\} \leq Pr\{\frac{1}{a_p}E[\lVert \mathbf{P}\bb \rVert_2^2]-\frac{1}{a_p}\lVert \mathbf{P}\bb \rVert_2^2 \geq \epsilon\}
\leq exp\Big(-\frac{2a_p^2\epsilon^2}{m\lVert \bb^2 \rVert_\infty^2}\Big)\\
\leq exp\Big(-\frac{2a_p^2p^2\epsilon^2}{mc^2}\Big).
\end{align*}  

Now in our scenario, assume $\B_a$ and $\B_b$ are two orthogonal spaces, i.e., $\B_a'\B_b = \mathbf{0}$. We set $\bb$ as $\frac{\bb_{ai}-\bb_{bj}}{\Vert \bb_{ai}-\bb_{bj} \Vert}$, where $\bb_{ai}$ and $\bb_{bj}$ are any column of $\B_a$ and $\B_b$ respectively. Then with probability $1-\delta$,
\begin{align*}
\frac{1}{a_1}\lVert \mathbf{P}(\bb_{ai}-\bb_{bj}) \rVert_2^2 \leq (1+\epsilon)\lVert \bb_{ai}-\bb_{bj} \rVert_2^2,\\
\frac{1}{a_p}\lVert \mathbf{P}(\bb_{ai}-\bb_{bj}) \rVert_2^2 \geq (1-\epsilon)\lVert \bb_{ai}-\bb_{bj} \rVert_2^2.
\end{align*}
With probability $1-2\delta$, we also have
\begin{align*}
a_p(\bb_{ai}^{\prime}\bb_{bj}-\epsilon\lVert \bb_{ai} \rVert_{2} \lVert \bb_{bj} \rVert_{2}) \leq \bb_{aiZ}^{\prime}\bb_{bjZ} \leq     a_1(\bb_{ai}^{\prime}\bb_{bj}+\epsilon\lVert \bb_{ai} \rVert_{2} \lVert \bb_{bj} \rVert_{2})
\end{align*}
From the foregoing two-side constraints, we can obtain that when $\frac{c^2}{2\epsilon^2}log(\frac{(k_a+k_b)^2}{\delta}) \leq m \leq \frac{2a_p^2p^2\epsilon^2}{c^2log\frac{(k_a+k_b)^2}{\delta}}$, $-a_p\epsilon \leq \bb_{aiZ}^{\prime}\bb_{bjZ}\leq a_1 \epsilon$ holds with probability $1-2\delta$, where $0 \leq a_p \leq a_1 \leq 1$. 
% \begin{equation*}
% a_p(1-\epsilon)\lVert \B \rVert_2^2\leq \lVert P\B \rVert_2^2 \leq a_1(1-\epsilon)\lVert \B \rVert_2^2
% \end{equation*}
So we can verify that the subspaces of $\B_a$ and $\B_b$ are approximately orthogonal when $m$ satisfies the foregoing conditions.
\section*{Appendix D: Proof of Theorem 1 and Theorem 2}
The simplified sampling procedure is to sample $Z$ by ranking $\Lambda_{(n+1)i}=\Big(2\hat{\X}_1^{\prime}\B_{ai}^{\prime}\B_{ai}\tilde{\bmu}_a-\bmu_a^{\prime}(\B_{ai}^{\prime}\B_{ai}\circ \bar{\mathbf{A}})\bmu_a\Big), i=1,...,p$ from the largest to the smallest and select the top $m$ variables. $\hat{\X}_1$ is generated by sampling $\hat{\btheta}_{a}$ from $\tilde{p}(\btheta_a,\mathbf{r})$, sampling $\hat{\mathbf{E}}$ from $N(\mathbf{0},\bSigma_{e})$ and getting $\hat{\X}_1=\B_a\hat{\btheta}_{a}+\hat{\mathbf{E}}$.

Since the posterior distribution of $\btheta_a$ is in spike-slab form, the distribution of $\Lambda_{(n+1)i}$ is Gaussian mixture distribution, which means it follows $2^{k_a}$ Gaussian distribution, each with different probability. If denote $S=\{1,2,...,k_a\}$, 
for any subset $S_0$ of $S$, we have
\begin{align*}
\Lambda_{i} \sim &N\Big(2\sum_{j \in S\setminus S_0}B_{aij}\mu_{aj}\sum_{k \in S}B_{aik}\mu_{ak}\alpha_{k}-\sum_{j \in S}B_{aij}^2\mu_{aj}^2\alpha_j-2\sum_{\forall j_1,j_2 \in S,j_1 \neq j_2}B_{aij_1}B_{aij_2}\mu_{aj_1}\mu_{aj_2}\alpha_{j_1}\alpha_{j_2},\\
&4(\sum_{j \in S\setminus S_0} B_{aij}^2s_j^2+\sum_{j \in S_0} B_{aij}^2vs_j^2+\sigma_e^2)(\sum_{j \in S} B_{aij}\mu_{aj}\alpha_j)^2\Big) 
\end{align*}
with probability $\prod_{j \in S \setminus S_0}\alpha_j\prod_{j \in S_0}(1-\alpha_j)$.

According to Theorem $5$ of \cite{wang2019frequentist}, the VB posterior converges to point mass of the true parameter value in distribution. Under our case, in normal condition, the true value of $\theta_{aj}$ equals $0$, $\forall j=1...k_a$. The posterior distribution that we obtain through VB method is in spike-slab form. For example, $q_j(\theta_{aj})\sim N(\mu_{aj}, s_j^2)$ with probability $\alpha_j$ and $q_j(\theta_{aj})\sim N(0, vs_j^2)$ with probability $1-\alpha_j$. Then suggested by Theorem $5$ of \cite{wang2019frequentist}, as $n \rightarrow \infty$,
\begin{align}
\label{eq:IC_q_property}
q_j(\theta_{aj}) \xrightarrow{d} \delta_0, \forall j,
\end{align}
where $\delta_0$ is a point mass at $0$. That suggests $\mu_{aj}\rightarrow 0$ and $s_j^2\rightarrow 0$. So in normal condition, $E(\Lambda_{(n+1)i})\rightarrow 0$ and $Var(\Lambda_{(n+1)i})\rightarrow 0$, $\forall j=1...k_a$, which means under the limit conditions, we sample the variables $Z(n+1)$ randomly.

Following a similar way, in abnormal condition, assume the anomaly relates to certain bases $\calA \subset S$. For $l \in \calA$, assume the anomaly relates to the $l^{th}$ base has change magnitude $\phi_l$. Then suggested by Theorem $5$ of \cite{wang2019frequentist}, as $n \rightarrow \infty$,
\begin{align}
\label{eq:OC_q_property}
&q_l(\theta_{al}) \xrightarrow{d} \delta_{\phi_l}, \forall l \in \calA \\
&q_j(\theta_{aj}) \xrightarrow{d} \delta_0, \forall j \in S-\calA
\end{align}
where $\delta_{\phi_l}$ is a point mass at $\phi_l$. That suggests $\mu_{al}\rightarrow \phi_l$, $\alpha_{l}\rightarrow 1$ and $s_l^2\rightarrow 0$.
The same as normal condition, $\mu_{aj}\rightarrow 0$ and $s_j^2\rightarrow 0$, $\forall j \neq l$. So in abnormal condition, $E(\Lambda_{(n+1)i})\rightarrow \sum_{l \in \calA}B_{ail}^2\phi_l^2+2\sum_{l_1,l_2\in \calA, l_1 \neq l_2}B_{ail_1}B_{ail_2}\phi_{l_1}\phi_{l_2}$ and $Var(\Lambda_{(n+1)i})\rightarrow 0$, $\forall j=1...k_a$. Similar proof can be extended to cases when anomaly relates to multiple bases. 

For general cases with $\btheta_{n}\B_{n}$, When $m \rightarrow \infty$ and $p \rightarrow \infty$ but the fraction $\frac{m}{p} \rightarrow \eta$, with $\eta$ being an arbitrary number between $0$ and $1$, according to the consistency of posterior estimation in Bayesian theory \citep{ghosh2007introduction}, $\tilde{\btheta}_{n} \rightarrow \btheta_n$. Then the properties of $\btheta_{a}$ in (\ref{eq:IC_q_property}) and (\ref{eq:OC_q_property}) still hold. Consequently, Theorem \ref{theorem:in control sampling} and Theorem \ref{theorem:out of control sampling} hold. 
\section*{Appendix E: Simulation Results for 1D and 2D Cases}
\label{Appendix5}
\begin{table}[h]
	\centering
	\caption{Average Detection Delays/ADDs(Standard Deviation of Detection Delays/STDs) for 1D data with $m=10$}
	
\begin{tabular}{cccccccc}
	\hline
	$\phi$& TRAS& CMAB(s) &NAS &SASAM &CDSSD &CDSSD(I) &ORACLE\\
	\hline
	0.0& 200(142)& 200(180) &200(221) &200(111) &200(266) &200(250) &200(361)\\
	0.1& 163(102)& 33.6(19.8) &192(212) &170(96.5) &\textbf{15.2}(18.0) &20.4(22.8) &2.90(2.60)\\
	0.2& 133(89.8)& 14.5(6.96) &161(176) &115(61.6) &\textbf{4.68}(3.67) &5.12(4.07) &1.30(0.55)\\
	0.3& 109(77.6)& 9.78(5.03) &151(166) &74.5(39.4) &2.85(2.22) &\textbf{2.84}(2.24) &1.05(0.24)\\
	0.4& 98.0(76.0)& 7.70(4.66) &136(148) &52.6(27.1) &2.26(1.63) &\textbf{2.13}(1.60) &1.00(0.08)\\
	0.5& 86.0(70.0)& 6.39(4.06) &141(157) &38.5(19.5) &2.00(1.50) &\textbf{1.73}(1.43) &1.00(0.00)\\
	0.6& 77.8(67.9)& 5.46(3.56) &135(151) &31.1(15.8) &1.77(1.31) &\textbf{1.64}(1.21) &1.00(0.00)\\
	0.7& 74.2(69.0)& 4.86(3.73) &130(143) &26.5(13.5) &1.63(1.07) &\textbf{1.54}(1.14) &1.00(0.00)\\
	0.8& 71.1(69.3)& 4.18(3.31) &135(149) &22.8(11.1) &1.59(1.16) &\textbf{1.48}(1.04) &1.00(0.00)\\
	0.9& 66.5(67.9)& 3.86(3.37) &125(140) &19.9(9.44) &1.54(1.07) &\textbf{1.46}(1.09) &1.00(0.00)\\
	1.0& 62.0(67.3)& 3.58(3.18) &124(142) &17.5(7.09) &1.45(1.01) &\textbf{1.41}(0.92) &1.00(0.00)\\
	\hline
\end{tabular}
\end{table}
\begin{table}[h]
	\centering
	\caption{Average Detection Delays/ADDs(Standard Deviation of Detection Delays/STDs) for 1D data with $m=20$}
	
	\begin{tabular}{cccccccc}
		\hline
		$\phi$& TRAS& CMAB(s) &NAS &SASAM &CDSSD &CDSSD(I) &ORACLE\\
		\hline
		0.0& 200(148)& 200(184) &200(548) &200(133) &200(292) &200(355) &200(361)\\
		0.1& 157(96.3)& 16.4(8.73) &161(460) &157(91.1) &\textbf{6.43}(7.00) &8.36(11.6) &2.90(2.60)\\
		0.2& 115(63.0)& 6.86(2.81) &151(415) &87.8(44.6) &\textbf{2.21}(1.91) &2.94(1.89) &1.30(0.55)\\
		0.3& 93.2(50.1)& 4.64(2.16) &125(371) &55.2(23.5) &\textbf{1.45}(0.93) &1.45(0.99) &1.05(0.24)\\
		0.4& 78.3(42.5)& 3.89(1.81) &111(344) &40.4(16.2) &1.23(0.69) &\textbf{1.17}(0.55) &1.00(0.08)\\
		0.5& 68.1(37.4)& 2.95(1.49) &107(323) &31.2(11.9) &1.16(0.61) &\textbf{1.12}(0.49) &1.00(0.00)\\
		0.6& 60.4(36.6)& 2.43(1.26) &93.0(299) &26.1(9.41) &1.15(0.63) &\textbf{1.09}(0.42) &1.00(0.00)\\
		0.7& 54.1(38.0)& 2.14(1.44) &76.1(259) &21.0(7.40) &1.14(0.49) &\textbf{1.07}(0.42) &1.00(0.00)\\
		0.8& 47.6(31.3)& 1.78(1.15) &87.3(289) &18.6(5.95) &1.11(0.45) &\textbf{1.05}(0.30) &1.00(0.00)\\
		0.9& 44.6(27.4)& 1.61(1.05) &87.8(292) &16.5(5.09) &1.09(0.38) &\textbf{1.04}(0.28) &1.00(0.00)\\
		1.0& 41.3(27.8)& 1.56(1.08) &75.8(253) &14.9(4.78) &1.09(0.45) &\textbf{1.05}(0.32) &1.00(0.00)\\
		\hline
	\end{tabular}
\end{table}
\begin{table}[h]
	\centering
	\caption{Average Detection Delays/ADDs(Standard Deviation of Detection Delays/STDs) for 1D data with $m=30$}
	
	\begin{tabular}{cccccccc}
		\hline
		$\phi$& TRAS& CMAB(s) &NAS &SASAM &CDSSD &CDSSD(I) &ORACLE\\
		\hline
		0.0& 200(145)& 200(180) &200(350) &200(134) &200(361) &200(444) &200(361)\\
		0.1& 158(97.8)& 10.5(4.13) &163(295) &137(78.5) &2.90(2.60) &\textbf{2.60}(3.11) &2.90(2.60)\\
		0.2& 114(52.0)& 4.31(1.22) &123(228) &74.0(33.6) &1.30(0.55) &\textbf{1.30}(0.67) &1.30(0.55)\\
		0.3& 90.1(36.3)& 2.79(0.67) &86.6(180) &47.2(17.9) &1.05(0.24) &\textbf{1.05}(0.23) &1.05(0.24)\\
		0.4& 74.5(27.60)& 2.12(0.35) &81.9(170) &34.8(12.4) &1.00(0.08) &\textbf{1.00}(0.08) &1.00(0.08)\\
		0.5& 63.9(23.1)& 1.93(0.32) &73.5(157) &26.7(8.65) &1.00(0.00) &\textbf{1.00}(0.03) &1.00(0.00)\\
		0.6& 55.5(20.4)& 1.57(0.50) &62.2(138) &22.1(6.80) &1.00(0.00) &\textbf{1.00}(0.00) &1.00(0.00)\\
		0.7& 50.0(16.1)& 1.20(0.40) &62.3(144) &18.9(5.40) &1.00(0.00) &\textbf{1.00}(0.00) &1.00(0.00)\\
		0.8& 45.2(0.33)& 1.09(0.28) &61.7(138) &16.6(4.95) &1.00(0.00) &\textbf{1.00}(0.00) &1.00(0.00)\\
		0.9& 41.0(12.5)& 1.03(0.16) &58.9(137) &14.5(3.96) &1.00(0.00) &\textbf{1.00}(0.00) &1.00(0.00)\\
		1.0& 38.6(11.9)& 1.00(0.05) &59.3(147) &13.1(3.74) &1.00(0.00) &\textbf{1.00}(0.00) &1.00(0.00)\\
		\hline
	\end{tabular}
\end{table}
\begin{table}[h]
	\centering
	\caption{Average Detection Delays/ADDs(Standard Deviation of Detection Delays/STDs) for 2D data with $m=20$}
	
	\begin{tabular}{cccccccc}
		\hline
		$\phi$& TRAS& CMAB(s) &NAS &SASAM &CDSSD &CDSSD(I) &ORACLE\\
		\hline
		0.0& 200(117)& 200(172) &200(173) &200(166) &200(201) &200(203) &200(479)\\
		0.1& 197(107)& 160(123) &203(181) &147(111) &\textbf{87.6}(90.8) &159(188) &1.71(1.57)\\
		0.2& 164(85.1)& 85.8(60.0) &193(176) &134(101) &\textbf{24.4}(17.5) &61.2(79.2) &1.09(0.34)\\
		0.3& 138(70.0)& 54.4(27.9) &202(187) &98.2(68.4) &\textbf{15.9}(12.7) &22.9(28.4) &1.02(0.15)\\
		0.4& 119(65.8)& 39.3(17.7) &184(176) &75.2(51.4) &12.9(10.7) &\textbf{12.1}(11.3) &1.00(0.08)\\
		0.5& 105(60.1)& 31.7(12.9) &192(174) &56.2(35.2) &11.2(9.96) &\textbf{8.67}(7.74) &1.00(0.00)\\
		0.6& 96.2(60.0)& 27.0(11.2) &187(182) &44.9(28.2) &10.2(9.94) &\textbf{6.23}(5.27) &1.00(0.00)\\
		0.7& 89.7(56.5)& 24.1(10.4) &201(188) &35.9(21.6) &9.45(8.93) &\textbf{5.59}(4.77) &1.00(0.00)\\
		0.8& 79.2(53.0)& 21.9(9.84) &191(185) &30.4(17.9) &9.02(9.26) &\textbf{4.81}(4.33) &1.00(0.00)\\
		0.9& 75.6(53.5)& 20.8(9.91) &196(188) &26.8(15.9) &8.22(9.24) &\textbf{4.38}(3.70) &1.00(0.00)\\
		1.0& 74.1(54.2)& 19.7(9.90) &192(177) &23.8(13.6) &7.47(8.07) &\textbf{3.90}(3.54) &1.00(0.00)\\
		\hline
	\end{tabular}
\end{table}
\begin{table}[h]
	\centering
	\caption{Average Detection Delays/ADDs(Standard Deviation of Detection Delays/STDs) for 2D data with $m=40$}
	
	\begin{tabular}{cccccccc}
		\hline
		$\phi$& TRAS& CMAB(s) &NAS &SASAM &CDSSD &CDSSD(I) &ORACLE\\
		\hline
		0.0& 200(116)& 200(171) &200(228) &200(161) &200(212) &200(285) &200(479)\\
		0.1& 193(107)& 114(93.3) &208(240) &139(102) &\textbf{31.9}(37.9) &145(156) &1.71(1.57)\\
		0.2& 152(72.0)& 46.7(27.2) &195(233) &119(92.1) &\textbf{10.5}(8.34) &36.2(51.4) &1.09(0.34)\\
		0.3& 125(59.5)& 30.0(10.7) &187(212) &88.5(60.4) &\textbf{6.68}(5.24) &12.6(11.8) &1.02(0.15)\\
		0.4& 104(51.0)& 23.6(7.59) &204(236) &66.0(45.7) &\textbf{5.33}(4.25) &6.47(5.75) &1.00(0.08)\\
		0.5& 91.8(47.9)& 20.4(7.29) &188(225) &50.5(30.6) &\textbf{4.74}(3.96) &4.74(3.55) &1.00(0.00)\\
		0.6& 81.6(44.7)& 18.4(7.24) &179(210) &39.9(24.6) &4.00(3.54) &\textbf{3.45}(2.74) &1.00(0.00)\\
		0.7& 73.8(42.0)& 16.3(7.14) &177(196) &30.5(17.0) &3.64(3.33) &\textbf{3.34}(2.59) &1.00(0.00)\\
		0.8& 66.1(40.8)& 15.7(7.28) &196(210) &26.7(14.5) &3.12(2.83) &\textbf{2.64}(1.81) &1.00(0.00)\\
		0.9& 59.8(37.4)& 15.3(7.29) &181(203) &22.4(11.1) &2.95(2.86) &\textbf{2.26}(1.66) &1.00(0.00)\\
		1.0& 56.3(37.5)& 14.1(7.35) &185(190) &19.6(9.62) &2.89(3.00) &\textbf{2.16}(1.54) &1.00(0.00)\\
		\hline
	\end{tabular}
\end{table}
\begin{table}[h]
	\centering
	\caption{Average Detection Delays/ADDs(Standard Deviation of Detection Delays/STDs) for 2D data with $m=60$}
	
	\begin{tabular}{cccccccc}
		\hline
		$\phi$& TRAS& CMAB(s) &NAS &SASAM &CDSSD &CDSSD(I) &ORACLE\\
		\hline
		0.0& 200(117)& 200(180) &200(238) &200(172) &200(251) &211(291) &200(479)\\
		0.1& 184(101)& 91.3(71.7) &188(234) &132(107) &\textbf{20.3}(25.5) &102(131) &1.71(1.57)\\
		0.2& 151(77.3)& 35.5(17.2) &196(244) &119(87.5) &\textbf{6.24}(4.90) &24.4(29.1) &1.09(0.34)\\
		0.3& 118(53.7)& 23.7(7.53) &187(241) &91.0(62.8) &\textbf{4.29}(3.06) &7.43(7.45) &1.02(0.15)\\
		0.4& 95.9(44.9)& 19.3(6.54) &168(203) &61.2(37.3) &\textbf{3.39}(2.47) &4.62(3.45) &1.00(0.08)\\
		0.5& 83.3(38.9)& 16.8(6.12) &199(246) &45.2(26.1) &\textbf{2.82}(2.05) &2.96(2.14) &1.00(0.00)\\
		0.6& 72.5(34.9)& 14.8(6.11) &191(251) &35.3(19.2) &2.57(2.32) &\textbf{2.48}(1.78) &1.00(0.00)\\
		0.7& 66.8(33.7)& 13.7(6.14) &190(243) &28.5(15.0) &\textbf{2.21}(1.58) &2.27(1.53) &1.00(0.00)\\
		0.8& 60.5(32.2)& 12.7(6.18) &178(232) &24.2(11.9) &1.92(1.35) &\textbf{1.90}(1.41) &1.00(0.00)\\
		0.9& 54.2(28.9)& 12.1(6.21) &168(213) &21.4(10.4) &1.91(1.43) &\textbf{1.64}(0.98) &1.00(0.00)\\
		1.0& 50.1(28.9)& 11.9(6.12) &168(217) &18.9(9.26) &1.90(1.48) &\textbf{1.54}(1.01) &1.00(0.00)\\
		\hline
	\end{tabular}
\end{table}
\end{document}